\newcounter{myctr}

\documentclass{ws-ijhr}

\PassOptionsToPackage{square, comma, sort&compress}{natbib}
\usepackage{natbib}				

\PassOptionsToPackage{fleqn}{amsmath} 
\usepackage{amsmath}
\usepackage{mathtools}

\usepackage{algorithm}
\usepackage{algpseudocode}
\numberwithin{algorithm}{section}
\renewcommand{\algorithmicrequire}{\textbf{Input: }}
\renewcommand{\algorithmicensure}{\textbf{Output: }}

\usepackage{tabu}
\usepackage{wrapfig}
\usepackage{subfig}

\usepackage[nolist]{acronym} 

\newcommand{\ie}{i.e.}

\newcommand{\sgn}{sgn}

\begin{document}

\markboth{Missura, Bennewitz, Behnke}{Capture Steps}

%
\catchline{}{}{}{}{}
%

\title{CAPTURE STEPS: \\ROBUST WALKING FOR HUMANOID ROBOTS}

\author{MARCELL MISSURA}

\address{Humanoid Robots Lab, Institute for Computer Science VI, University of Bonn,
\\Endenicher Allee 19A, 53115 Bonn, Germany\\
missura@cs.uni-bonn.de}

\author{MAREN BENNEWITZ}

\address{Humanoid Robots Lab, Institute for Computer Science VI, University of Bonn,
\\Endenicher Allee 19A, 53115 Bonn, Germany\\
maren@cs.uni-bonn.de}

\author{SVEN BEHNKE}

\address{Autonomous Intelligent Systems, Institute for Computer Science VI, University of Bonn,
\\Endenicher Allee 19A, 53115 Bonn, Germany\\
behnke@cs.uni-bonn.de}

\maketitle

\begin{history}
\received{Day Month Year} %
\revised{Day Month Year}  %
\accepted{Day Month Year} %
\end{history}

\begin{abstract}

Stable bipedal walking is a key prerequisite for humanoid robots to reach their
potential of being versatile helpers in our everyday environments. Bipedal
walking is, however, a complex motion that requires the coordination of many
degrees of freedom while it is also inherently unstable and sensitive to
disturbances. The balance of a walking biped has to be constantly maintained.
The most effective way of controlling balance are well timed and placed recovery
steps---capture steps---that absorb the expense momentum gained from a push or a
stumble. We present a bipedal gait generation framework that utilizes step
timing and foot placement techniques in order to recover the balance of a biped
even after strong disturbances. Our framework modifies
the next footstep location instantly when responding to a disturbance and generates
controllable omnidirectional walking using only very little
sensing and computational power. We exploit the open-loop stability of a
central pattern generated gait to fit a linear inverted pendulum model to
the observed center of mass trajectory. Then, we use
the fitted model to predict suitable footstep locations and timings in order to
maintain balance while following a target walking velocity. Our
experiments show qualitative and statistical evidence of one of the strongest 
push-recovery capabilities among humanoid robots to date.

\end{abstract}


\begin{acronym}
        \acro{UML}{Unified Modeling Language}
        \acro{CoM}{Center of Mass}
        \acro{CoP}{Center of Pressure}
        \acro{CPG}{Central Pattern Generator}
        \acro{LIPM}{Linear Inverted Pendulum Model}
        \acro{TDA}{Trunk Deviation Angle}
        \acro{IMU}{Inertial Measurement Unit}
        \acro{ZMP}{Zero Moment Point}
\end{acronym}

\section{Introduction}

Bipedal walking is an energy efficient and versatile means of locomotion suitable
for covering large distances in a wide variety of terrains. Its principle dynamics
can be likened to an inverted pendulum that is constantly falling and requires 
active control in order to remain balanced---for example by stepping into
the right place at the right time. Undoubtedly, it would be of great benefit if we
were able to replicate a walking controller with human-like
capabilities. Unfortunately, as easy as walking comes to us humans, 
the design of 
walking controllers for bipedal robots has proven to be rather challenging.
The widespread state of the art covers basic walking on flat surfaces in the
absence of disturbances. Push recovery, walking on rough terrain, and agile
footstep control are active research topics. The dominant strategy to make a
robot walk is to abstract from the complex body and to represent its centroidal
momentum with the inverted pendulum model. In most cases, the mathematically
tractable Linear Inverted Pendulum Model is used to deduce controllers that
steer and balance the pendulum in a way that the \acl{ZMP}---the assumed pivot
point of the inverted pendulum---stays within the boundaries of preset footsteps
that have been planned ahead. The trajectory of the
point mass model is then transformed into a whole-body walking motion by
tracking the pendulum motion with the pelvis, connecting the pendulum base
locations with smooth swing foot trajectories in Cartesian space, and computing
the resulting motor commands using inverse kinematics. This approach works, but 
has not yet achieved the versatility and robustness of the human gait.

The bipedal walk generation technique presented here differs from the state of
the art in a number of aspects. Instead of designing a low-dimensional model
and forcing a robot to follow its motion, we first craft a central
pattern-generated~(CPG) whole-body motion that can produce an open-loop stable
gait. A
low-dimensional inverted pendulum model is then fitted to match the observed
center of mass trajectory of the open-loop motion. The fitted model can then be
used to predict the state of balance at the end of the step, and to compute 
the location and the timing of a footstep that is expected to restore balance
towards a stable limit cycle. Our approach augments the CPG gait with balance control
by modifying
timing and landing coordinates of footsteps in a non-intrusive way, leaving
the execution of the stepping motions up to the underlying pattern generator.
The result is a robust and controllable omnidirectional walk that does not
derogate the natural dynamics by forcing the center of mass onto a
plane---a consequence when a low-dimensional model is imposed on the robot as many other approaches do.

Our algorithm requires very little computational power and only basic sensory equipment.
Inertial sensors in the torso are used to estimate its attitude, and joint
position sensors are used to reconstruct the pose of the robot. No force or torque
sensors are required. A precise robot model is not required either, as masses,
torques, and forces are not involved in our computations. A rough kinematic model
describing approximate link lengths suffices. We are
also able to relax precision requirements on the actuation level. We operate our
robot in a compliant setting with low-gain position-controlled actuators. 
Despite its low requirements, it achieves one of the
strongest push recovery capabilities among
humanoid robots to date.

\section{Related Work}
\label{chap:relatedwork}

\ac{ZMP} preview control~\citep{Kajita03} is the most popular approach to
bipedal walking. A number of pre-planned footsteps are
used to define a future \ac{ZMP} reference trajectory. A continuous \ac{CoM}
trajectory that minimizes the \ac{ZMP} tracking error, the jerk of the \ac{CoM},
and the deviation from terminal conditions at the end of the preview horizon, is
then generated by solving a quadratic program \citep{Wieber}. The optimization is computationally expensive,
but can be performed in real time. In theory, once a smooth and stable model is
computed, a robot closely following the motion of the model should be stable,
too. By using the \ac{ZMP} preview control scheme, high quality robots
\citep{HRP4C,HUBO} can walk on flat ground as long as disturbances are
small. More advanced gait controllers from the \ac{ZMP} preview family
\citep{WieberWithSteps,HRPPush,Stephens3} also consider foot placement in
addition to \ac{ZMP} control by including the footstep locations in the
optimization process.

A sampling-based \ac{ZMP} preview controller that includes 
adaptive foot-placement has been proposed by \citet{Urata}.
Instead of optimizing the \ac{CoM} trajectory for a single \ac{ZMP} reference, a
fast sampling method is used to generate a whole set of lower quality
\ac{ZMP}/\ac{CoM} trajectory pairs for three steps into the future.
Triggered by a disturbance, the algorithm selects and executes the best available motion
according to given optimization criteria. Resampling during execution of the
motion plan is not possible. This method was demonstrated to
produce push recovery capabilities on a real robot. Highly specialized hardware
was used to meet the speed and precision requirements.

Another interesting planning approach was presented by \citet{Honda} where walking, running, and hopping are planned in parallel as alternative modes of locomotion. The best plan is selected according to stability and energy consumption criteria. Footstep locations and timing are computed by a gradient decent algorithm during walking.

The capture point \citep{Pratt:CapturePoint} is an appealing indicator of balance.
It describes the location on the ground where a biped would 
need to step in order to come to a complete stop.
\citet{OttCaptureStep} proposed the use of a capture point trajectory as a
reference input for gait generation, instead of the \ac{ZMP}. The capture
point approach is much simpler and faster to compute than \ac{ZMP} preview
control. A capture point based preview controller was demonstrated on Toro
\citep{Toro} and on the Atlas robot to produce a walk of the same quality as the 
classic \ac{ZMP} preview controller.

A drawback of all of the aforementioned approaches is that the motion of a
low-dimensional model is computed first,
and then the robot is forced to follow its trajectory as closely as possible.
This imposes precise position tracking requirements on the hardware in order to
preserve the stability predicted by the model. Furthermore, a low-dimensional
model strongly simplifies the walking motion by design. Following the model closely
results in an unnatural, plane-restricted motion of the pelvis---typically with
extensive use of bent knees.

The inverse approach of starting with the motion before balance originates
from passive
dynamic walking pioneered by \citet{McGeer}. His experiments
proved that the passive dynamics of legs with freewheeling joints is
sufficiently stable to walk down a shallow slope. With a minimal amount of
actuation to restore lost energy, passive walking on level ground is also
possible \citep{MIKE,Denise,Cornell}. The graceful motions of these bipedal
constructions strongly resemble the human walk and suggest that the core
principle of biological gaits may also be passive dynamics with minimal control
effort. 

CPG walking adds actuation, but no control of
balance. However, a small basin of attraction around the upright pose allows for controllable, open-loop stable walking. Interestingly, in the competitive environment of RoboCup, where humanoid robots play soccer, CPG walking is
the dominant approach. Perhaps the most advanced RoboCup gait was presented for the Nao standard
platform by \citet{BHuman}. Based on the solution of a system of linear pendulum
equations, the timing and trajectory of the pendulum motion is adjusted
online in order to land the swing foot as closely as possible to a desired
step size.

The DARwIn-OP platform \citep{DARwInrobot} comes
with a fast and reliable walk that has been described by \citet{SJ2}. The core
gait has a strong similarity with \ac{ZMP} preview control. Future
footstep locations are placed in a queue as reference. However, instead of the expensive
\ac{CoM} trajectory optimization that includes jerk minimization, the \ac{CoM}
trajectory is generated open-loop and in closed form using simple \acl{LIPM}
equations that do not limit the jerk.

By modeling virtual forces that keep the robot upright and pull it in the
desired direction of locomotion, \citet{VMC} created the Virtual Model Control
approach. The virtual forces are mapped to torques of the actuators such that
the same trunk motion is produced as the forces would. With this method, the
two-dimensional robot Spring Flamingo showed a fluid and
natural looking walk that was robust enough to reject small disturbances and to
walk up and down on slopes.

The work presented here has its origins in \citep{Missura:LateralCaptureSteps},
where first lateral stability was investigated with the conclusion that controling
the step timing is effective at recovering the lateral oscillation after a push
from the side. Then, the concept was extended to the sagittal direction in
simulation \citep{Missura:OmnidirectionalCaptureSteps} and implemented on a real
robot \citep{Missura:WalkingWithCaptureSteps}. Onboard learning of the sagittal step
size has also been investigated \citep{Missura:SagittalLearningRobot} to a degree
where a robot was able to learn to absorb a strong push after only a few failed
steps.

\section{Capture Step Framework}
\label{chap:overview}
\acresetall

\begin{figure}[h] 
\centering 
\includegraphics[height=7cm]{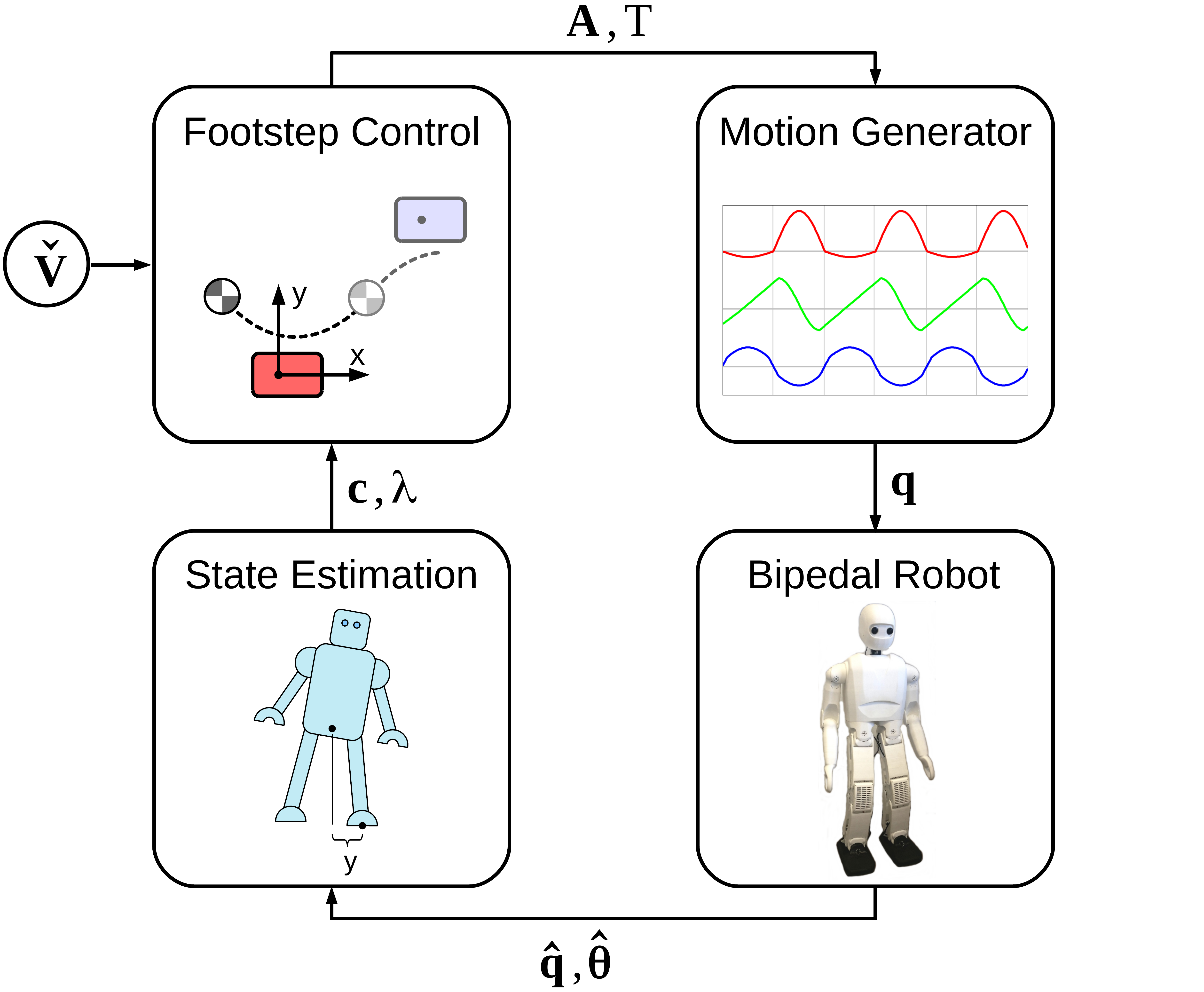}
\caption[Overview of the Capture Step Framework]{Overview of the Capture Step
Framework. The State Estimation component (bottom left) reconstructs the tilted pose of
the robot using the measured joint angles $\boldsymbol{\hat{q}}$ and the torso inclination 
$\boldsymbol{\hat{\theta}}$. From the reconstructed pose, the state of the center 
of mass $\boldsymbol{c}$ and the support foot
indicator $\lambda$ are gained. The Footstep Control module (top
left) uses the center of mass state $\boldsymbol{c}$ and the support foot
indicator $\lambda$ to compute the swing amplitude $\boldsymbol{A}$ and the timing $T$
of the next footstep in order to track the desired walking velocity
$\boldsymbol{\check{V}}$ while maintaining balance. The Motion
Generator (top right) executes a timed whole-body stepping motion with the
commanded step size and generates the joint position targets $\boldsymbol{q}$.}
\label{fig:overview}
\end{figure}

The bipedal gait generation method presented here is called \emph{Capture
Step Framework}. Figure~\ref{fig:overview} illustrates the components of the
framework organized in the circular layout of a control loop. The robot itself
is part of the loop. It receives motor targets from the control
software and provides sensor data about its internal state. The three main software
components are: State Estimation, Footstep Control, and Motion
Generation.

The input into the control loop is a target velocity parameter $\boldsymbol{\check{V}}
\in \lbrack-1,1\rbrack^{3}$ that controls the sagittal, the lateral, and the
rotational velocity of the gait. The target velocity determines
the size of the steps the robot should produce. The framework attempts to realize
the commanded velocity as good as it can, but it may deviate from it in order to
maintain balance. Alternatively, it is possible to use a reference step size as
input instead of the velocity in order to command the robot to follow a footstep
plan \citep{Missura:Thesis}.

The Motion Generator is a CPG that generates whole-body stepping motions
composed of superimposed leg-lifting and leg-swinging motion primitives. The
motion primitives are parameterized in a way that the footstep location at the
end of the step can be controlled by the swing amplitude parameter
$\boldsymbol{A} \in \mathbb{R}^{3}$ in the sagittal, lateral, and rotational directions.
The timing of the steps is controlled by the step time parameter $T$. In the end, the whole-body motion
trajectory is conveniently expressed as a signal of joint angles
$\boldsymbol{q}$ that are passed on to the robot. The robot tracks the joint
angles using PD-controlled servo motors.

Using the joint angles $\boldsymbol{\hat{q}}$ and the torso inclination
$\boldsymbol{\hat{\theta}}$ as measured by the sensors of the robot, the State
Estimation module reconstructs the tilted whole-body pose. From the
reconstructed pose, the motion of a fixed point on the body frame is tracked and
used as a low-dimensional representation of balance. The coordinates and
velocities of this fixed point---hereafter referred to as the \acf{CoM} state
$\boldsymbol{c} = \left(c_x, \dot{c}_x, c_y, \dot{c}_y\right)$---are determined
in the sagittal (x, forward) and the lateral (y, sideward) directions with
respect to the coordinate frame of the support foot with the sign $\lambda \in
\{-1, 1\}$. We assign -1 to the left foot and 1 to the right foot. 

The \ac{CoM}
state vector $\boldsymbol{c}$, the sign $\lambda$ of the support foot, and the
desired velocity $\boldsymbol{\check{V}}$, are the inputs into the Footstep
Control module, where a \ac{LIPM} is used to compute the swing amplitude 
$\boldsymbol{A}$ and the time $T$ of the next footstep in order to
keep the center of mass balanced while obeying the desired velocity
$\boldsymbol{\check{V}}$ as closely as possible. The swing amplitude
$\boldsymbol{A} \in \mathbb{R}^{3}$ and the step time $T$ become the inputs 
of the Motion Generator module.

One iteration of the control software loop requires 0.12\,ms to compute on a single 1.3\,GHz core and thus can be operated with a high frequency. We are using an update frequency of 100\,Hz.

In the following, we introduce the modules of the Capture Step Framework shown in 
Figure~\ref{fig:overview} in detail.

\section{Motion Generator}
\label{chap:cpg}

The nonzero size of the support polygon of a humanoid robot allows for some 
passive stability that can be exploited to implement stable walking with 
an open-loop CPG. The \mbox{NimbRo} CPG gait was
originally proposed by \citet{Behnke06} and extended by \citet{Missura:NimbRoGait} and
\citet{Missura:Thesis}. Using oscillating motion patterns for the legs and the arms, the
CPG generates an omnidirectional gait that allows a humanoid robot to step in the
sagittal, lateral, and rotational directions. The step sizes in all three directions, and
the step timing, can be modified quickly and independently during walking, which gives 
rise to a relatively agile and controllable gait.

\subsection{Abstract Kinematic Interface} 
\label{chap:leginterface}

\begin{figure}[t]
\centering
\includegraphics[height=5cm]{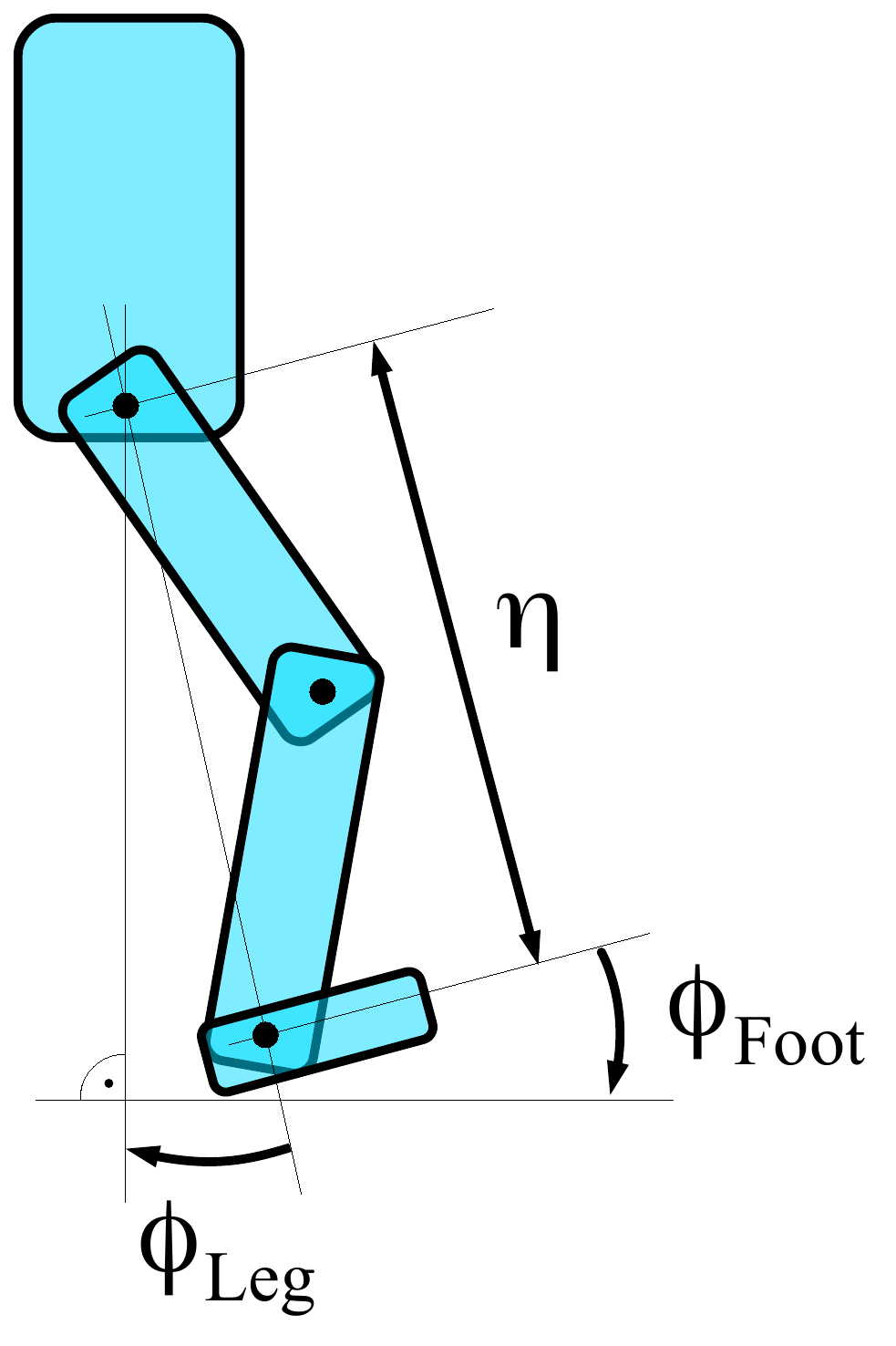}\label{fig:leginterfacea}
\caption[The Leg Interface]{The Leg Interface encapsulates leg pose
control with three abstract parameters: the leg extension $\eta$, the leg angle
$\boldsymbol{\phi}_{Leg}$, and the foot angle $\boldsymbol{\phi}_{Foot}$.}
\label{fig:leginterface}
\end{figure}

The motion patterns of the CPG are embedded in the parameter space of a kinematic
abstraction layer that we named \emph{Leg Interface}. The Leg Interface exhibits three
abstract parameters to control the pose of a leg---the leg extension $\eta$, the leg angle
\mbox{$\boldsymbol{\phi}_{Leg}$}, and the foot angle \mbox{$\boldsymbol{\phi}_{Foot}$}.
The meaning of the parameters is illustrated in Figure~\ref{fig:leginterface}.
The leg extension $\eta \in \lbrack0,1\rbrack$ allows the leg to be extended and retracted like a prismatic
joint. The leg angle \mbox{$\boldsymbol{\phi}_{Leg} =
(\phi_{Leg}^{Roll}, \phi_{Leg}^{Pitch}, \phi_{Leg}^{Yaw})$} determines the
rotation of the leg with respect to the trunk in the
roll, pitch, and yaw directions. The foot angle
parameter \mbox{$\boldsymbol{\phi}_{Foot} = \left(\phi_{Foot}^{Roll},
\phi_{Foot}^{Pitch}\right)$} is used to determine the rotation of the foot with
respect to the trunk. Formally, the Leg Interface is a function
$
\left(\boldsymbol{\phi}_{Hip},\ \phi_{Knee},\ \boldsymbol{\phi}_{Ankle}\right)
= \mathcal{L}\left(\eta,\ \boldsymbol{\phi}_{Leg},\
\boldsymbol{\phi}_{Foot}\right)
$
that encapsulates the mapping of the
abstract parameters to joint angles
\mbox{$\boldsymbol{\phi}_{Hip} = (\phi_{Hip}^{Roll},
\phi_{Hip}^{Pitch}, \phi_{Hip}^{Yaw})$} for the hip joint, $\phi_{Knee}$ for the
knee joint, and \mbox{$\boldsymbol{\phi}_{Ankle}
= (\phi_{Ankle}^{Roll}, \phi_{Ankle}^{Pitch})$} for the ankle joint using the equations
\begin{align}
\begin{bmatrix} 
{\phi^{\prime \,}}_{Leg}^{Pitch} \\
{\phi^{\prime \,}}_{Leg}^{Roll} 
\end{bmatrix} 
&= 
R(-\phi_{Leg}^{Yaw})
\begin{bmatrix} 
\phi_{Leg}^{Pitch} \\ 
\phi_{Leg}^{Roll} 
\end{bmatrix},\label{eq:rot}\\ 
\zeta &= \arccos(1 - \eta),\label{eq:legex}\\ 
\left(\phi_{Hip}^{Yaw}, \phi_{Hip}^{Roll}, \phi_{Hip}^{Pitch}\right) &= \left(\phi_{Leg}^{Yaw}, {\phi^{\prime\,}}_{Leg}^{Roll}, {\phi^{\prime\,}}_{Leg}^{Pitch} - \zeta\right)\\ 
\phi_{Knee} &= 2\zeta,\\ 
\left(\phi_{Ankle}^{Pitch}, \phi_{Ankle}^{Roll}\right) &= \left( \phi_{Foot}^{Pitch} - {\phi^{\prime \,}}_{Leg}^{Pitch} -\zeta, \phi_{Foot}^{Roll} - {\phi^{\prime\,}}_{Leg}^{Roll} \right),
\label{eq:leginterface} 
\end{align}
where $R(-\phi_{Leg}^{Yaw})$ is a rotation by the negative leg yaw. Note that the motion abstraction layer is
essentially model free. Unlike for inverse kinematics, the actual sizes of the body segments 
do not need to be known.

The parameter space of the Leg Interface offers an intuitive way to encode 
motion components that a robot would naturally perform during walking. For example,
lifting the leg at the beginning of the swing phase, and stretching it 
shortly before the heel strike, can be achieved using the leg extension
parameter $\eta$. Swinging the leg to the front and back is achieved by
modulating the pitch angle parameter $\phi_{Leg}^{Pitch}$ with an oscillating signal.

\subsection{CPG Gait}

\begin{figure}[h] 
\centering 
\includegraphics[width=0.7\textwidth]{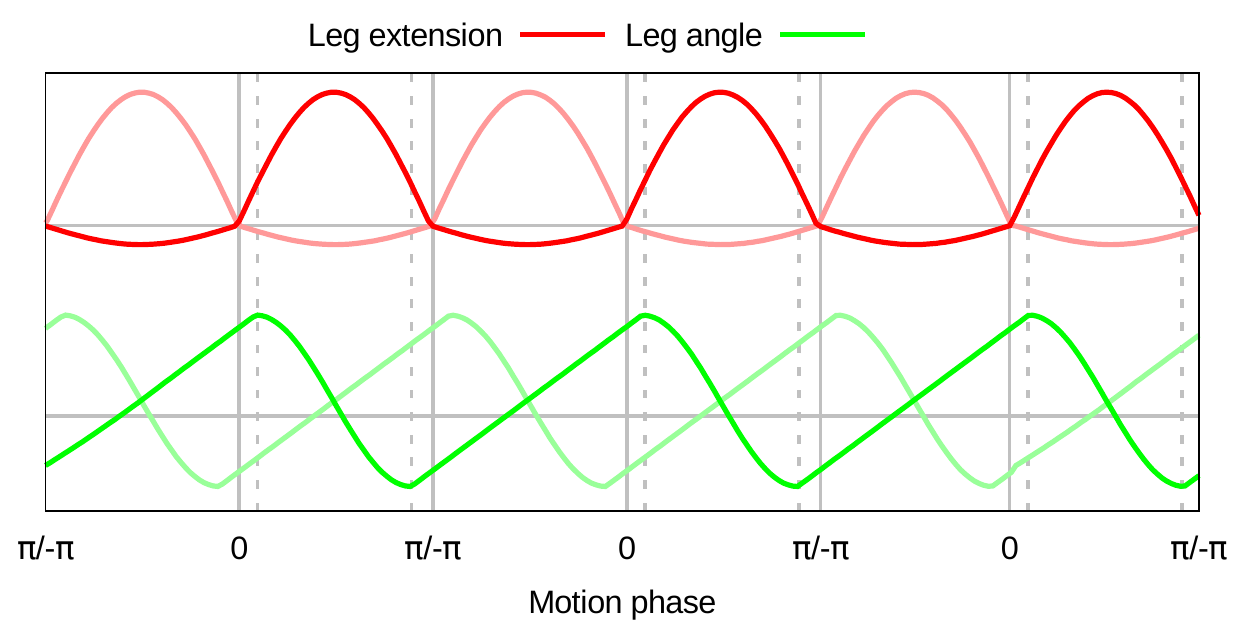}
\caption[The Motion Patterns of the Gait Generator]{The main ingredients
of the gait motion are rhythmical leg lifting (top) and a leg swing motion
(bottom). The solid vertical lines indicate
the expected times of support exchange. The dashed vertical lines indicate the
swing start and swing end timings. The patterns for the left leg shown in
faint color are phase shifted by $\pi$.}
\label{fig:motionpattern} 
\end{figure}

The walking motion generated by the CPG can be subdivided into motion
primitives that produce Leg Interface parameters. The final motion
pattern is composed as the sum of the outputs of all motion primitives. The most
important motion primitives for leg lifting and leg swinging are shown in Figure~\ref{fig:motionpattern} and will be introduced
shortly. For a complete list of all involved motion primitives, we refer to
\citep{Missura:Thesis}.

The oscillation of the motion signal is driven by a motion phase \mbox{$\mu
\in [-\pi,\pi)$}. The motion phase is incremented in every iteration 
of the main control loop \mbox{$\mu_{t+1} = \mu_t + \delta_\mu$}, where the increment
$
\delta_\mu = \frac{\rho\nu}{T}
$
is computed from the remaining motion phase
\begin{equation} 
\nu = 
\begin{cases}
-\mu, &\mbox{if } \mu \leq 0 \\
\pi - \mu, &\mbox{otherwise},
\end{cases}
\end{equation} 
the commanded step time $T$, and the main control loop iteration period $\rho = 0.01$\,s. 

The amplitude of the leg lifting and leg swinging patterns is
determined by the swing amplitude vector \mbox{$\boldsymbol{A}=\left(A_x, A_y,
A_{\psi}\right)$} with parameters for the roll, pitch, and yaw
directions. The swing amplitude and the step time are computed by the Footstep Control module 
that will be presented in detail in Section~\ref{chap:footstepcontrol}.

\subsubsection{Leg Lifting} 

The leg lifting primitive is an alternating shortening of the legs. As shown
in Figure~\ref{fig:motionpattern} on the top, the leg lifting primitive activates
the leg extension parameter with a sinusoidal function
\begin{equation} 
\eta(\mu,\boldsymbol{A}) = 
\begin{cases}
\sin(\mu)\,\left(K_{1} + K_{3}\,||\boldsymbol{A}||_\infty\right), &\mbox{if } \mu \leq 0 \\
\sin(\mu)\,\left(K_{2} + K_{4}\,||\boldsymbol{A}||_\infty\right), &\mbox{otherwise}
\end{cases}
\label{eq:leglift}
\end{equation} 
that depends on the motion phase \mbox{$\mu \in [-\pi, \pi)$} and the swing
amplitude $\boldsymbol{A}$. 
Notably, the leg lifting primitive makes a
distinction between a support phase---when the motion phase \mbox{$\mu \leq 0$}
and the foot is on the ground---and a swing phase---when the motion phase
\mbox{$\mu > 0$} and the foot is in the air. During the support phase, a small
push is applied against the ground. During the swing phase, the foot is lifted
up into the air and can be swung. The configuration parameters $K_{1}$ and
$K_{2}$ describe the push height during the support phase and the step height
during the swing phase, respectively. The configuration variables $K_{3}$ and 
$K_{4}$ intensify the push and the lift depending on the $L_\infty$ norm of 
the swing vector $\boldsymbol{A}$, \ie, the foot is lifted higher 
the faster the robot is walking. The support exchange is expected to occur at 
motion phases $\mu = 0$ and $\mu = \pm\pi$.

The same function is used to generate the motion for both legs by computing
$\eta(\mu_r,\boldsymbol{A})$ for the right leg with $\mu_r = \mu$ and $\eta(\mu_l,\boldsymbol{A})$
for the left leg with a phase shifted
\begin{equation} 
\mu_l = 
\begin{cases}
\mu+\pi, &\mbox{if } \mu \leq 0 \\
\mu-\pi, &\mbox{otherwise}.
\end{cases}
\end{equation}

\subsubsection{Leg Swing} 

To swing the leg in any direction, we use the leg swing pattern
shown on the bottom of Figure~\ref{fig:motionpattern}. 
The leg
is swung forwards with a sinusoidal motion and pushed backwards with a linear
motion during its support phase.
The forward swing is not perfectly embedded into the motion phase. Swing
phase configuration parameters $K_{\mu_0}$ and $K_{\mu_1}$ are used to delay the
start of the swing and to rush the finish of the swing to happen earlier than the
nominal support exchange at motion phase $\mu = \pm\pi$. 
Essentially, the shortened swing accounts for an implicit double support time where the weight
of the robot shifts from one leg to the other.

To generate the leg swing pattern, we first compute a motion phase
dependent unit swing oscillator
\begin{equation}
\zeta(\mu) = 
\begin{cases}
\frac{2\left(\mu+2\pi - K_{\mu_1}\right)}{2\pi - K_{\mu_1} +
K_{\mu_0}} -1 , & \mbox{if } -\pi \leq \mu < K_{\mu_0}\\
\cos\left(\frac{\pi\left(\mu - K_{\mu_0}\right)}{K_{\mu_1} - K_{\mu_0}}\right),
& \mbox{if } K_{\mu_0} \leq \mu < K_{\mu_1} \\ 
\frac{2\left(\mu -
K_{\mu_1}\right)}{2\pi - K_{\mu_1} + K_{\mu_0}} - 1, & \mbox{if } K_{\mu_1} \leq
\mu < \pi, 
\end{cases}
\end{equation}
which incorporates
the sinusoidal forward swing, the linear swing during the support phase, and the
swing timing parameters $K_{\mu_0}$ and $K_{\mu_1}$. Then, we use the swing
vector $\boldsymbol{A}$ to modulate the amplitude of the unit swing
oscillator in the roll, pitch, and yaw directions, and compute the leg angle
parameters with the equations
\begin{align}
\phi_{Leg}^{Roll}(\lambda, \nu, \boldsymbol{A}) &= -\zeta(\nu) A_x K_{5} -
\lambda \max\{|A_x|K_{6},\,|A_{\psi}| K_{7}\},\label{eq:legswingroll}\\
\phi_{Leg}^{Pitch}(\lambda, \nu, \boldsymbol{A}) &= \zeta(\nu) A_y K_{8},\label{eq:legswingpitch}\\ 
\phi_{Leg}^{Yaw}(\lambda, \nu, \boldsymbol{A}) &= \zeta(\nu) \, A_{\psi} \,
K_{9} - \lambda |A_{\psi}| K_{10}. \label{eq:legswingyaw}
\end{align}
where $\lambda \in \{-1, 1\}$ denotes the sign of the leg (left or right) the pattern is generated for. 
The leg swing equations (\ref{eq:legswingroll}-\ref{eq:legswingyaw}) differ in
the three directions. In the pitch direction, the legs swing fully from front to back. 
The maximum swing amplitude for forward walking
and backward walking is configured using the step size parameter $K_{8}$.
In the roll direction, however, the legs would collide. Therefore, leg roll angle offsets
$K_{6}$ and $K_{7}$ are added proportionally to the roll and yaw swing
amplitude $A_x$ and $A_{\psi}$, causing the legs to spread out when
walking in the lateral direction, and when the robot is turning. In the yaw
direction, an amplitude-dependent yaw angle offset can be configured using the
parameter $K_{10}$. 
As with the leg lifting, the swing pattern is computed for both legs with 
the same function. 

\subsubsection{Arm Motion}

The arm motion is generated in an analogous fashion. Similar to the Leg
Interface, an Arm Interface provides an abstract actuator space where the
length of the arm and the angle of the arm can be manipulated independently.
The arms are swung with the same swing pattern as the legs,
but antagonistically to the legs, \ie, the right arm swings forward when the
left leg does, and the left arm swings forward when the right leg does. The 
role of the arm motion is to counteract the rotation about the vertical axis
that would otherwise be induced by the inertia of the swinging leg.

\subsection{CPG Gait Performance}



The demonstration video\footnote{https://youtu.be/HESyHEPNdd8} shows a
number of different humanoid robots that this CPG walk 
has been used on. All of these robots walked well on a flat floor and 
demonstrated outstanding performance in RoboCup soccer games.


\section{State Estimation}
\label{chap:stateestimation}

The State Estimation module (shown on the bottom left in
Figure~\ref{fig:overview}) reconstructs the tilted whole-body pose of the robot
for the purpose of extracting the \ac{CoM} state $\boldsymbol{c}$ and the
support leg sign $\lambda \in [-1, 1]$. To this end, it uses a kinematic model
and sensory information obtained from the robot.  The obtained sensor values are
the joint angles  $\boldsymbol{\hat{q}}$ as measured by motor encoders, and the
angle $\boldsymbol{\hat{\theta}}$ of the trunk as reported by the IMU.

\subsection{Tilted Whole-Body Pose Reconstruction}
\label{chap:posereconstruction}

\begin{figure}[h]
\centering
\includegraphics[height=6cm]{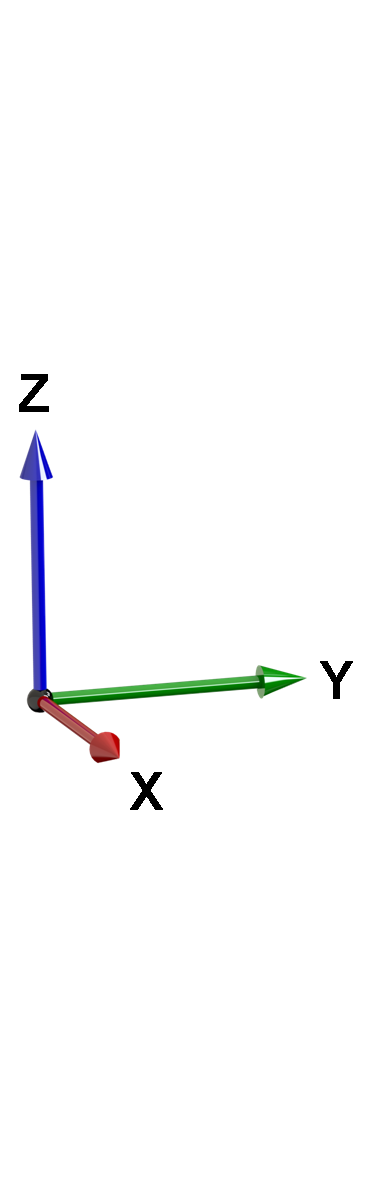}
\hspace{0.8cm}
\includegraphics[height=6cm]{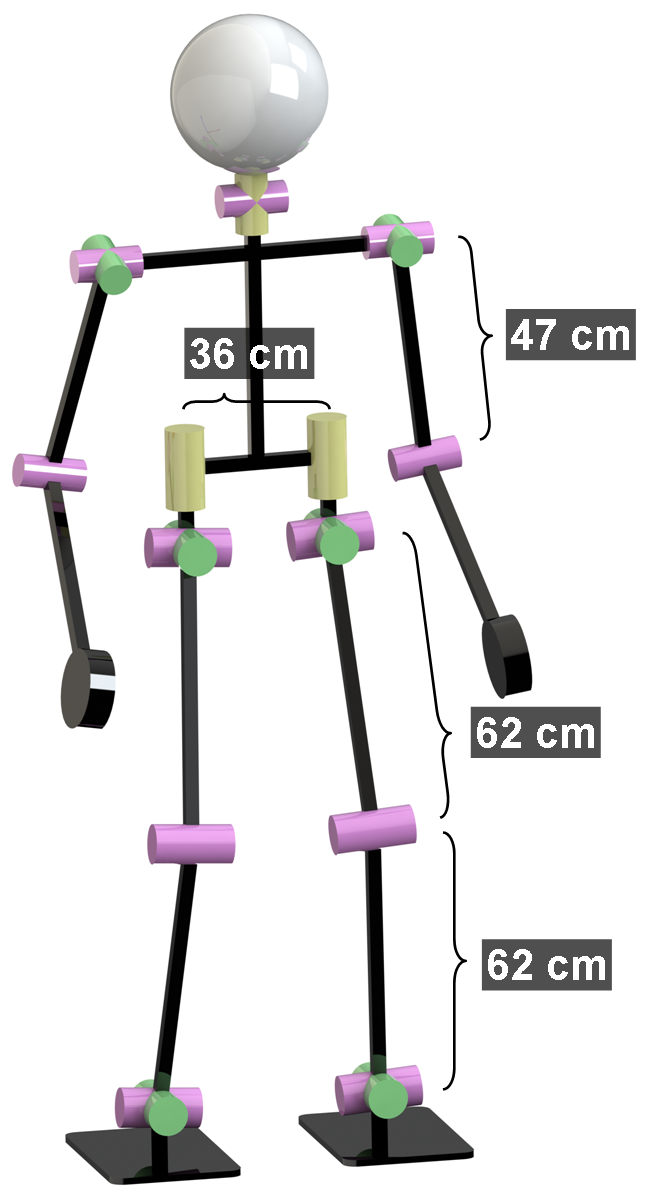}
\hspace{0.8cm}
\includegraphics[height=6cm]{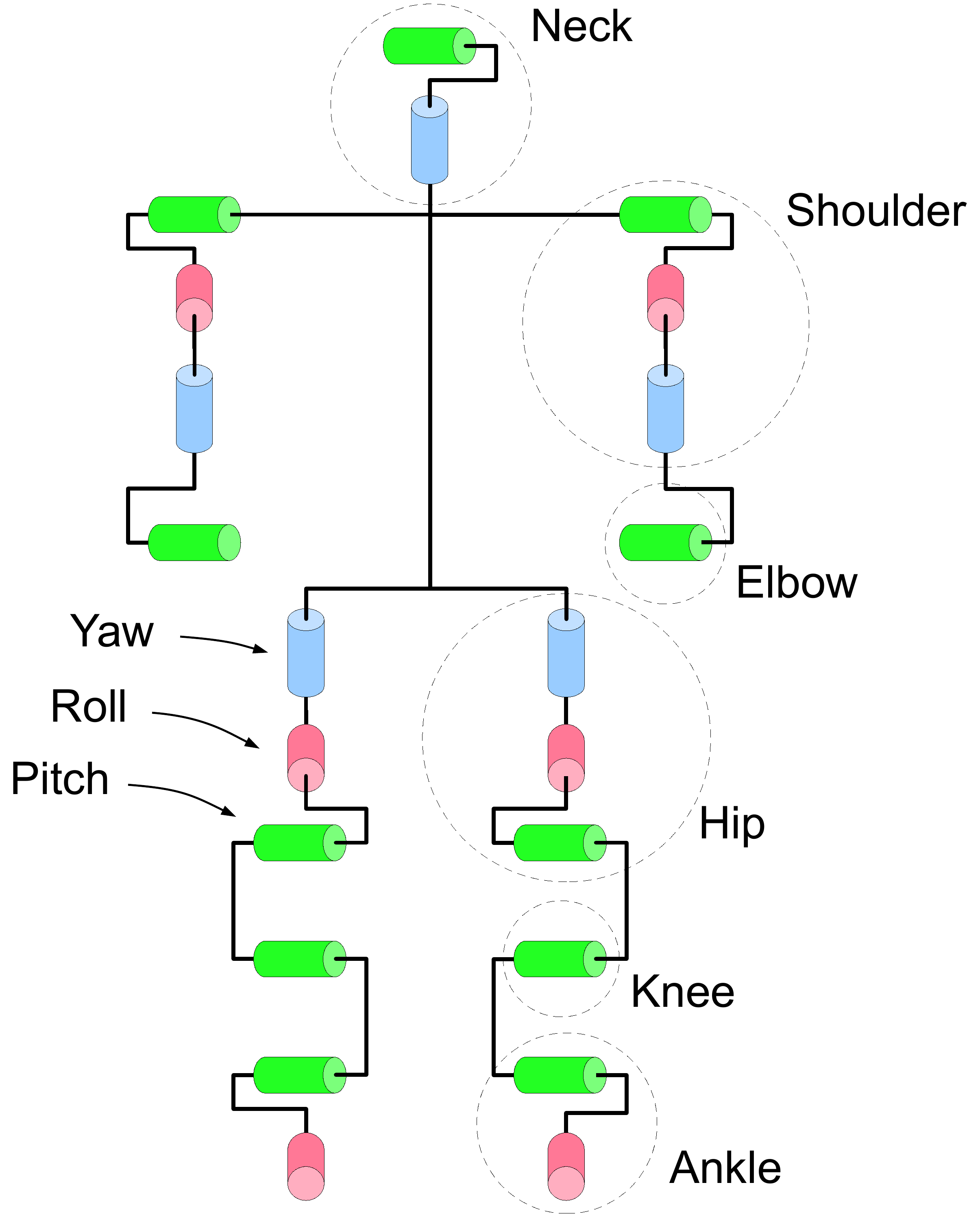}
\caption[The Kinematic Chain]{A generic kinematic chain that applies to most
humanoid robots.}
\label{fig:kinematicmodel}
\end{figure}

The kinematic model we use for the pose reconstruction is illustrated in
Figure~\ref{fig:kinematicmodel}.  It is a generic humanoid kinematic chain that
we use for different robots. The
lengths of the links of the skeleton were not adjusted to the measurements of the
specific robot we used in our experiments since the absolute values for the 
lengths do not matter, only their relation to each other.

For the pose reconstruction, the measured joint angles $\boldsymbol{\hat{q}}$
are applied to set the kinematic model in pose with a forward kinematics
algorithm. Once in pose, the entire kinematic model is rotated around the center
of the current support foot such that the trunk attitude equals the roll and
pitch angles $\boldsymbol{\hat{\theta}} = (\hat{\theta}_{roll},
\hat{\theta}_{pitch})$ measured from the robot. We consciously neglected the
fact that the robot would rotate about one of the edges of the support foot, and
not about the center of the foot. This eliminates the need to determine the edge
to rotate about, and disposes of a potential source of jitter at the cost of a
negligible error.

\subsection{Center of Mass State Estimation}
\label{chap:comstateestimation}

Using the reconstructed pose, we extract the
position of the center point in between the hip joints 
with respect to a footstep frame. The footstep frame is set to the ground projection 
of the new support foot in the moment of a detected support exchange. Support exchange 
detection is discussed in Section~\ref{chap:supportlegestimation}. The ground projected
support frame is horizontally aligned with the floor, but preserves 
the yaw orientation of the new support foot.
During the step, the footstep frame remains fixed until the next support
exchange occurs. With respect to the footstep frame, we compute the coordinates
of the ground projected center point between
the hip joints and obtain the \ac{CoM} state $\boldsymbol{c} = \left(c_x,
\dot{c}_x, c_y, \dot{c}_y\right)$. The values of
the derivatives $\dot{c}_x$ and $\dot{c}_y$ have to be determined by numerical
differentiation and are hence prone to noise.

\subsection{Support Foot Estimation}
\label{chap:supportlegestimation}

The support foot estimation is a continuous process that can be initialized with
either the right or the left foot. If after the pose reconstruction outlined
above the vertical coordinate of the swing foot has a value lower than the
vertical coordinate of the support foot, the roles of the feet are switched and
the sign $\lambda \in \{-1,1\}$ of the support foot is set to either $\lambda
= -1$ for the left foot, or $\lambda = 1$ for the right foot. In this moment, the
support frame is relocated to the ground projection of the new support foot. In
order to avoid erratic changes of the support foot sign when both feet are on
the ground, after every change of the support role, we require the vertical
distance between the feet to exceed $5$\,mm before another support exchange is
allowed to occur. Note that this support foot detection method is
based on the assumption that the floor is horizontal and flat, and at least one
foot touches the ground at all times. Based on these assumptions, support foot
detection is possible without foot pressure or ankle torque sensors, but the
method does not scale to non-planar surfaces.

\begin{figure}[t] 
\includegraphics[width=0.49\textwidth]{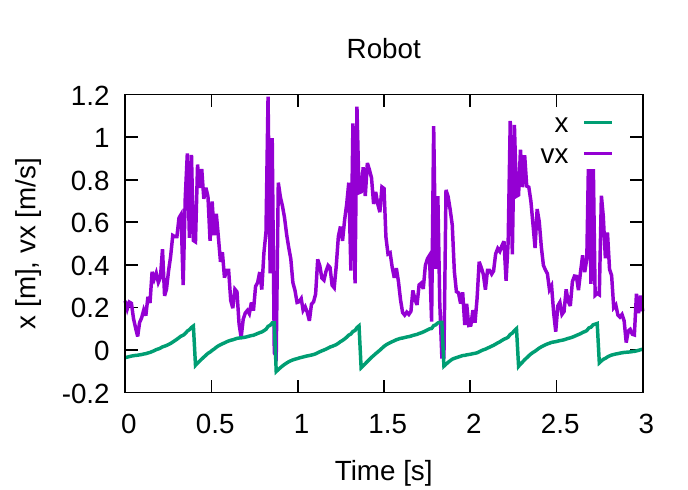}
\includegraphics[width=0.49\textwidth]{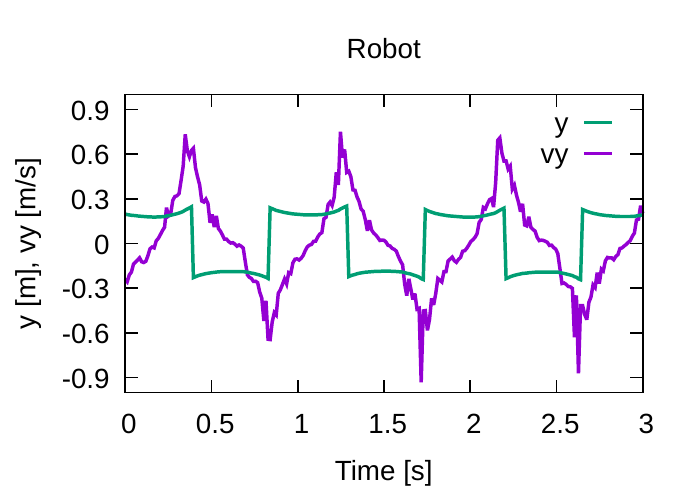}
\includegraphics[width=0.49\textwidth]{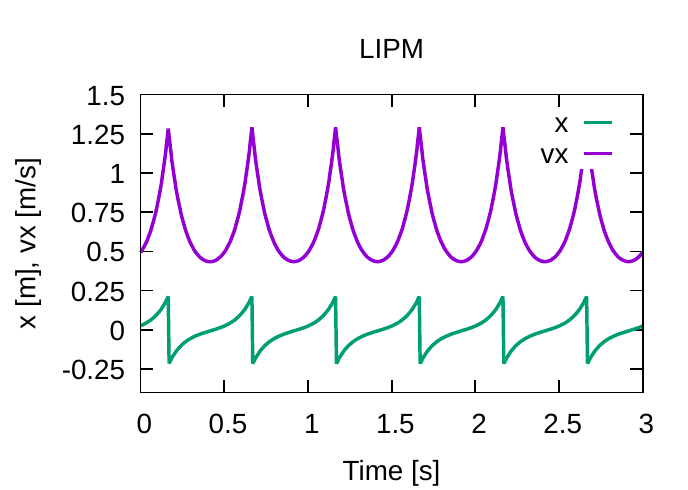}
\includegraphics[width=0.49\textwidth]{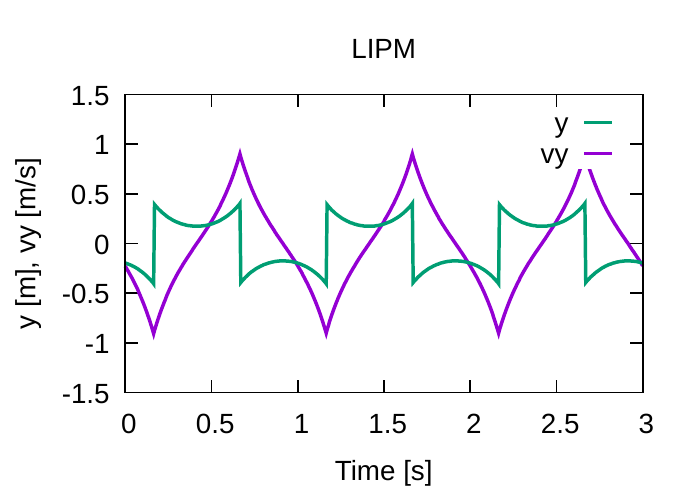}
\caption[Center of Mass Motion Data]{\acl{CoM} data in the sagittal (left column) 
and the lateral (right column) directions. In the top row, real robot data is shown. 
In the bottom row, the trajectory of a simulated \acl{LIPM} is shown. A strong 
resemblance between the real robot and the simulated model can be seen.}
\label{fig:comdata}
\end{figure}

\subsection{Experimental Validation}

Figure~\ref{fig:comdata} shows \ac{CoM} positions and velocities in the sagittal
and lateral directions as estimated with the tilted kinematic pose
reconstruction method during walking. Real robot data is shown in the top row,
simulated data is shown in the bottom row. It is obvious that the velocity
estimates are rather noisy, especially in the moment of the support exchange.
The Capture Step controller includes a predictive filter presented in
Section~\ref{chap:predictivefilter} for the smoothing of the \ac{CoM} state.
Most importantly, the \ac{CoM} data of the real robot resembles the motion of
the simulated pendulum in both the sagittal
and the lateral directions.

\section{Footstep Control}
\label{chap:footstepcontrol}

The Footstep Control module (shown on the top left in Figure~\ref{fig:overview})
implements balance-preserving gait control functions using the \ac{CoM} state
$\boldsymbol{c} = \left(c_x, \dot{c}_x, c_y, \dot{c}_y\right)$ and the support
leg sign $\lambda \in \{-1,1\}$. The outputs of the Footstep Control are the
swing amplitude $\boldsymbol{A} = \left(A_x, A_y, A_{\psi}\right)$ and the
step time $T$---the remaining time until the next support exchange. In its core,
the Footstep Control generates a reference trajectory for the \ac{CoM}. The
reference trajectory is the limit cycle that the \ac{CoM} would follow under
perfect conditions when executing stepping motions with the commanded velocity
$\boldsymbol{\check{V}}$. The reference trajectory is gained by having the robot walk with
the open-loop CPG gait and approximating the observed \ac{CoM} trajectory with a
parameterized \acl{LIPM}. Robust and controllable walking performance is
achieved by driving the \ac{CoM} towards the reference trajectory by means of 
\ac{ZMP}, step timing, and foot placement control
strategies.

\begin{algorithm}[t]
\caption{Footstep Control}
\label{alg:footstepcontrol}
\algorithmicrequire{Desired velocity $\boldsymbol{\check{V}}$}
\Comment{Command input}\\ \algorithmicrequire{CoM state $\boldsymbol{c}$},
support foot $\lambda$ \Comment{From the State Estimation}\\
\algorithmicensure{Step parameters $(\boldsymbol{A}, T)$}
\Comment{Swing amplitude and timing} \begin{algorithmic}[1]
\Function{FootstepControl}{$\boldsymbol{\check{V}}, \boldsymbol{c}, \lambda$}
\State $\boldsymbol{s}$ $\gets$ \Call{ReferenceTrajectory}{$\boldsymbol{\check{V}}, \lambda$} 
\State $(\boldsymbol{c}, \lambda)$ $\gets$ \Call{PredictiveFilter}{$\boldsymbol{c}, \lambda$} 
\State $(\boldsymbol{Z}, \boldsymbol{A}, T)$ $\gets$
\Call{BalanceControl}{$\boldsymbol{s}, \boldsymbol{c}, \lambda$} \State \Return $(\boldsymbol{A}, T)$
\EndFunction
\end{algorithmic}
\end{algorithm}

The footstep control
function is given in Algorithm~\ref{alg:footstepcontrol}.
Three main computation steps can be identified within 
\textsc{FootstepControl($\boldsymbol{\check V}$, $\boldsymbol{c}$, $\lambda$)}:
\textsc{ReferenceTrajectory($\boldsymbol{\check{V}}$, $\lambda$)} 
computes a target state $\boldsymbol{s}$ that represents the ideal \ac{CoM} trajectory, 
\textsc{PredictiveFilter($\boldsymbol{c}$, $\lambda$)} smooths the \ac{CoM} state input and overcomes latency by
means of prediction, and \textsc{BalanceControl($\boldsymbol{s}$,
$\boldsymbol{c}$, $\lambda$)} computes the \ac{ZMP} $\boldsymbol{Z}$,
the swing amplitude $\boldsymbol{A}$, and the step time $T$, which drive 
the \ac{CoM} state $\boldsymbol{c}$ towards the target state $\boldsymbol{s}$. 
Note that only the swing amplitude $\boldsymbol{A}$ and the step timing $T$ 
are returned by  \textsc{FootstepControl($\boldsymbol{\check V}$, $\boldsymbol{c}$, $\lambda$)} since the CPG does not need the
\ac{ZMP} for the generation of the stepping motion.

In the following, we first introduce the \acl{LIPM}---the mathematical
model that represents the principle dynamics of bipedal walking. After that, 
we discuss each step of Algorithm~\ref{alg:footstepcontrol} in detail.

\subsection{Linear Inverted Pendulum Model}
\label{chap:lip}

\subsubsection{One-dimensional Model}

\begin{figure}[t]
\centering
	\subfloat[1D]
	{\includegraphics[height=3.5cm]{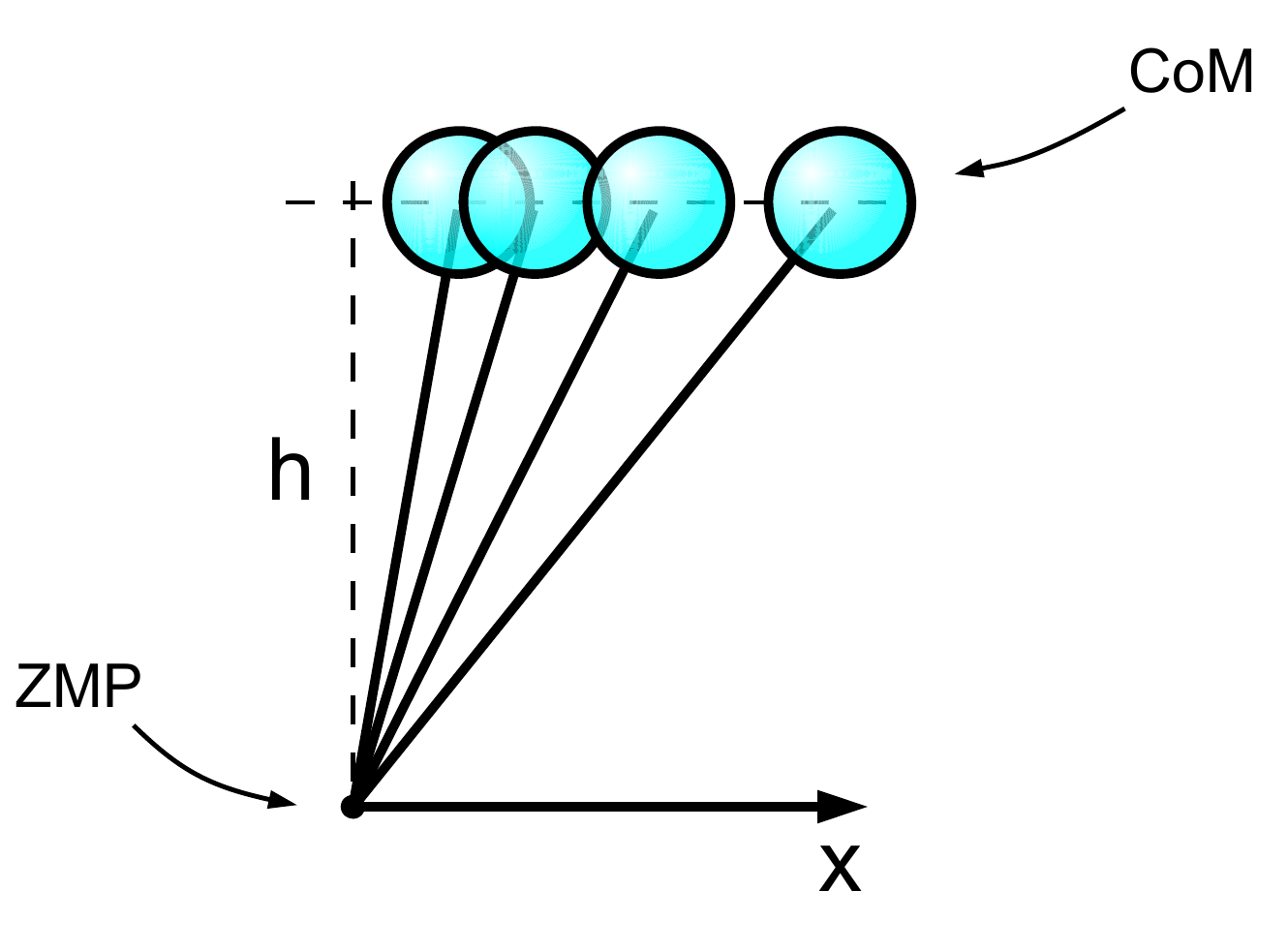}\label{fig:lip1d}}
	\hspace{1cm}
	\subfloat[2D with ZMP offset]
	{\includegraphics[height=3.5cm]{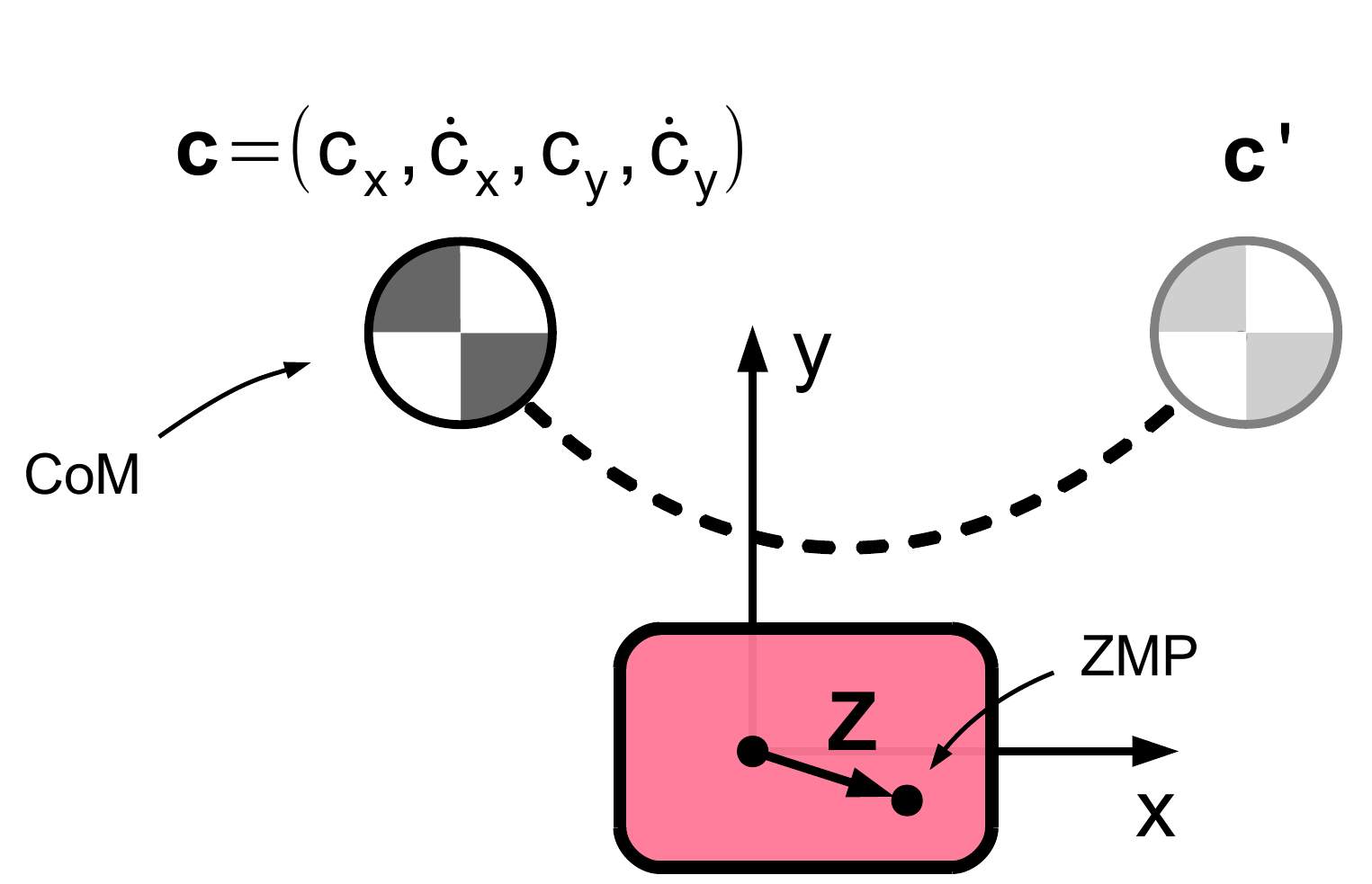}\label{fig:lip2d}}
\caption[The Linear Inverted Pendulum Model]{(a) The one-dimensional \acl{LIPM}
with pendulum height $h$ and \acl{CoM} coordinate $x$. (b) Using the orthogonal
superposition of two \aclp{LIPM}, the \acl{CoM} motion is approximated in a
two-dimensional plane. A small \acl{ZMP} offset
$\boldsymbol{Z}$ can be used for limited influence on the motion of the
\acl{CoM} $\boldsymbol{c}$.}
\label{fig:lip}
\end{figure}

The \acf{LIPM} was originally proposed by \citet{LIMP}.  It is a linearized
version of an inverted pendulum and resembles a bipedal walker standing on one
support leg, falling away from the pendulum base. Figure~\ref{fig:lip1d}
illustrates a one-dimensional \ac{LIPM}. The quantity of interest is the
horizontal displacement $x$ of the \ac{CoM} with respect to the pendulum base. The
pendulum base, the pivot point of the pendulum, the \ac{ZMP}, and the \ac{CoP}
are all different names for the same concept. 

The \ac{LIPM} describes the motion of the \ac{CoM} using the differential
equation 
\begin{equation} 
\ddot{x} = C^2x 
\label{eq:lip} 
\end{equation} 
for some
constant $C$. Typically, a value of $C = \sqrt{g/h}$ is
used where $g = 9.81$\,m/s$^2$ is the gravitational constant and $h$ is
the assumed constant height of the center of mass. The \ac{LIPM} suffers from inaccuracies due to its oversimplicity and should be corrected \citep{balgauer}. In our approach, we identify
the value of $C$ experimentally to fit the \ac{LIPM} as closely as possible to
the observed behavior of an individual robot.

The simple \ac{LIPM} differential equation (\ref{eq:lip})
has a closed form solution. For an initial state $\left(x_0,
\dot{x}_0\right)$, the location and the velocity of the future state at time
$t$ are computed by
\begin{align}
x(t, x_0, \dot{x}_0) &= x_0 \cosh(C t) + \frac{\dot{x}_0}{C}\sinh(C t)\label{eq:lipstatex},\\
\dot{x}(t, x_0, \dot{x}_0) &= x_0 C \sinh(C t)+\dot{x}_0\cosh(C t).\label{eq:lipstatevx} 
\end{align}
The time $t$ when the \ac{CoM}
reaches a future location $x$, or velocity $\dot{x}$, is given by
\begin{align}
t_{pos}(x, x_0, \dot{x}_0) &= \frac{1}{C} \ln{\left(\frac{x}{c_1} \pm
\sqrt{\frac{x^2}{c_1^2} - \frac{c_2}{c_1}}\right)} \label{eq:liptimex}, \\ 
t_{vel}(\dot{x}, x_0, \dot{x}_0) &= \frac{1}{C} \ln{\left(\frac{\dot{x}}{c_1 C} \pm
\sqrt{\frac{\dot{x}^2}{c_1^2 C^2} + \frac{c_2}{c_1}}\right)}
\label{eq:liptimevx},
\end{align}
where
$c_1 = x_0 + \frac{\dot{x}_0}{C}$ and $c_2 = x_0 - \frac{\dot{x}_0}{C}$.
Unless the pendulum is
disturbed by external forces, the orbital energy
\begin{equation}
E(x, \dot{x}) = \frac{1}{2}\left(\dot{x}^2 - C^2x^2\right)
\label{eq:lipenergy}
\end{equation}
remains constant along a trajectory. 

\subsubsection{Two-dimensional Model with ZMP}
\label{chap:lip2d}

We model a two-dimensional \ac{CoM} motion using
two uncoupled \ac{LIPM} equations
\begin{equation}
\begin{bmatrix}
  \ddot{x} \\
  \ddot{y}
\end{bmatrix}
=
\begin{bmatrix}
  C^2 & 0 \\
  0 & C^2
\end{bmatrix}
\begin{bmatrix}
  x \\
  y 
\end{bmatrix}.
\label{eq:lipm2d}
\end{equation}
The $x$ dimension describes the sagittal motion and the $y$ dimension describes
the lateral motion. Additionally, we extend this passive model with \ac{ZMP}
control as illustrated in Figure~\ref{fig:lip2d}. The origin of the coordinate
frame is located underneath the ankle joint of the support foot. A small
\ac{ZMP} offset $\boldsymbol{Z} = (Z_x, Z_y)$ can relocate the pendulum base
within the foot and influence the future motion of the \ac{CoM}. If we assume
that the \ac{ZMP} offset remains constant, we can incorporate the \ac{ZMP}
offset $\boldsymbol{Z}$ into the equations (\ref{eq:lipstatex}) and
(\ref{eq:lipstatevx}) in a trivial manner and formulate a two-dimensional
\ac{LIPM} predictor function \mbox{$\boldsymbol{c^{\prime}}$ = \textsc{LipmPredict($\boldsymbol{c}$, $\boldsymbol{Z}$, $t$)}} that
predicts the future \ac{CoM} state $\boldsymbol{c^{\prime}}$ at time $t$, given
the current \ac{CoM} state $\boldsymbol{c}$ at time $t = 0$ and the \ac{ZMP}
offset $\boldsymbol{Z}$, with the equations
\begin{align}
c^{\prime}_x &= x(t,\ c_x - Z_x,\ \dot{c}_x) + Z_x,\nonumber\\
\dot{c}^{\prime}_x &= \dot{x}(t,\ c_x - Z_x,\ \dot{c}_x),\nonumber\\
c^{\prime}_y &= x(t,\ y_x - Z_y,\ \dot{c}_y) + Z_y,\nonumber \\
\dot{c}^{\prime}_y &= \dot{x}(t,\ c_y - Z_y,\ \dot{c}_x).
\label{eq:limppredict}
\end{align}

\subsection{Reference Trajectory Generation}
\label{chap:referencetrajectory}

\begin{figure}[h] 
\centering
\subfloat[Sagittal Motion]{\includegraphics[width=0.48\textwidth]{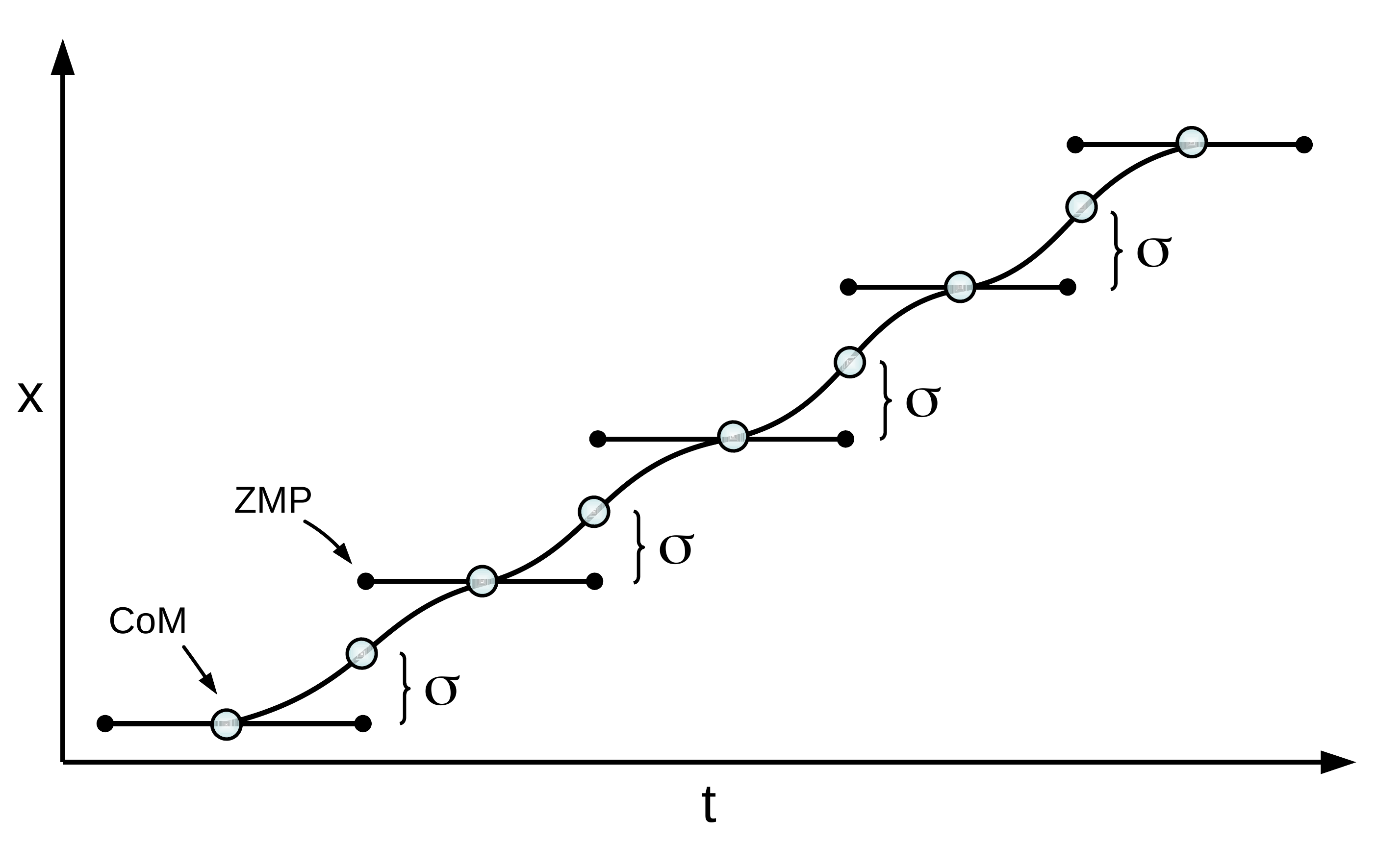}}
\subfloat[Lateral Motion]{\includegraphics[width=0.48\textwidth]{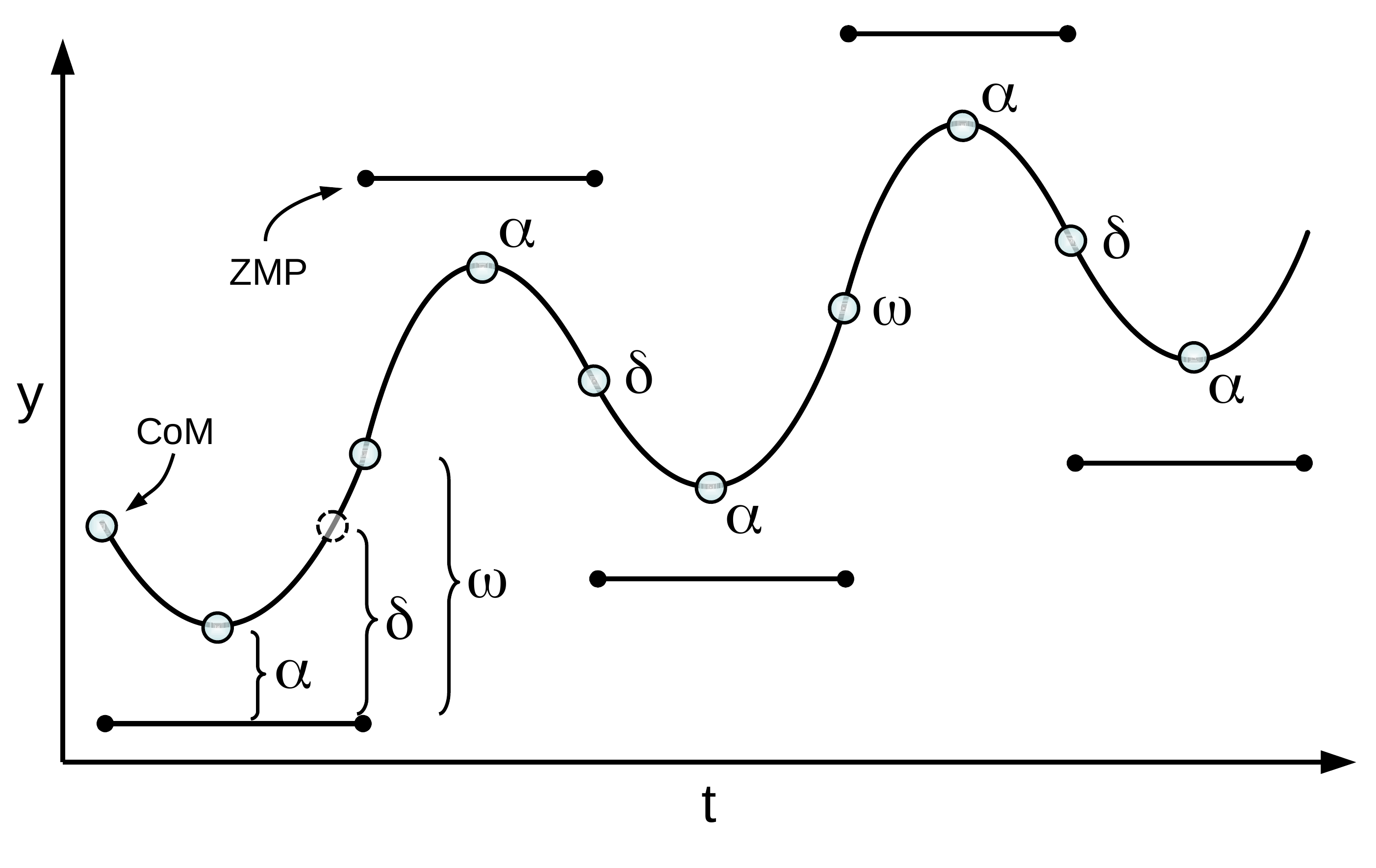}}
\caption[The Two-Dimensional Reference Trajectory]{The
\acl{CoM} reference trajectory is composed of (a) a sagittal
motion and (b) a lateral motion. Four configuration parameters define the
maximum sagittal \acl{CoM} displacement $\sigma$, the lateral apex distance
$\alpha$, and the minimal and maximal support exchange locations $\delta$ and
$\omega$. The support exchange is modeled as an
instantaneous relocation of the pendulum base such that in the moment of the
support exchange, the \acl{CoM} is in the center between
the pivot points.}
\label{fig:parameters}
\end{figure}

The gait generation cycle leans on the computation of a nominal trajectory that
describes the ideal motion of the \ac{CoM}. The shape of the nominal trajectory
is determined by the desired walking velocity $\boldsymbol{\check{V}}$, 
the pendulum constant $C$ (Eq.~\ref{eq:lip}), and configuration parameters named
$\alpha$, $\delta$, $\omega$, and $\sigma$. It does not
depend on the current state of the \ac{CoM}. 

Figure~\ref{fig:parameters} 
illustrates the schematics of the nominal trajectory and the meaning of the
parameters. In the sagittal direction, the point mass crosses the base of the 
pendulum once
in every step. The parameter $\sigma$ defines the displacement of the \ac{CoM}
with respect to the foot when walking forward with the maximum velocity
$\check{V}_x=1$. The \ac{CoM} displacement is zero when the robot is walking in
place and negative when the robot walks backwards. 

In the lateral direction, the 
point mass oscillates between two supports and
never crosses the base of the pendulum. The distance between the pivot point and
the apex of the trajectory is denoted by $\alpha$. The lateral \ac{CoM} velocity is
zero in this point. The support exchange occurs when the lateral \ac{CoM}
location is within a range bounded by $\delta$ and $\omega$. When walking in
place, the support exchange occurs at distance $\delta$. When
walking with a non-zero lateral velocity, the walker first takes a long step
with the leading leg and the support exchange occurs at a location up to the
upper bound $\omega$ where the lateral walking velocity $\check{V}_y=1$. The 
leading step is followed by a shorter trailing step where the support exchange 
always occurs at distance $\delta$, independent of the size of the leading step.
The values of the reference trajectory parameters $\alpha$, $\delta$, $\omega$,
and $\sigma$, and the pendulum constant $C$, can be determined using data 
collected from a robot walking with
the open-loop CPG.

The pendulum base is assumed to stay stationary  during a step, and to instantly
relocate in the moment of the support exchange in a way that the position of the
\ac{CoM} at the end of the step is in the center between the old and the new pendulum
bases. A double support phase is not included in our nominal trajectory model.

Assuming that in the ideal case the motion of the \ac{CoM} follows the laws of the \ac{LIPM} on a
constant-energy orbit, a single state \mbox{$\boldsymbol{s} = (s_x, \dot{s}_x,
s_y, \dot{s}_y)$} is sufficient to represent the nominal trajectory. We choose
the nominal state $\boldsymbol{s}$ to be the end-of-step \ac{CoM} state, \ie, 
the \ac{CoM} state in the moment of the next support exchange. 
The nominal lateral support exchange location is
\begin{equation}
\xi_y =  \begin{cases} \lambda
\left(\delta+|\check{V}_y|\left(\omega-\delta\right)\right), & \mbox{if }
\lambda = \sgn(\check{V}_y) \\ \lambda \delta, & \mbox{otherwise}. \end{cases}
\end{equation} 
We differentiate between the leading step case, where the lateral support
exchange location $\xi_y$ is between $\delta$ and $\omega$, and the trailing
step case, where the lateral support exchange always occurs at distance
$\delta$. The sagittal support exchange location is trivially given by 
\begin{equation}
\xi_x = \check{V}_x \sigma. 
\end{equation}
The complete nominal support exchange state can now be computed as 
\begin{equation} 
\boldsymbol{s} = 
\begin{bmatrix} 
s_x\\
\dot{s}_x\\ 
s_y\\ \dot{s}_y 
\end{bmatrix} 
= 
\begin{bmatrix} 
\xi_x\\ 
C\,\xi_x\, \mathrm{csch}\left(C \tau\right)\\  
\xi_y\\ 
\lambda C \sqrt{\xi_y^2 - \alpha^2}
\end{bmatrix}. 
\label{eq:nominalstate} 
\end{equation} 
$\tau = t_{pos}(\xi_y, \lambda\alpha, 0)$ (Eq.~\ref{eq:liptimex}) 
is thereby the time it takes for the \ac{CoM} to travel from the lateral apex $\alpha$
to the lateral support exchange coordinate $\xi_y$.
The nominal state
$\boldsymbol{s}$ is expressed in coordinates relative to the current support
foot. The computation of the nominal state $\boldsymbol{s}$ using Eq.
(\ref{eq:nominalstate}) is equivalent to the
$\boldsymbol{s}$ = \textsc{ReferenceTrajectory}($\boldsymbol{\check{V}}, \lambda$)
computation step in line 2 of the Footstep Control algorithm~\ref{alg:footstepcontrol}.

Along with the nominal support exchange state, we also compute the nominal step time
$\check{T}$, the time we expect a step to take in the ideal case. We set 
\mbox{$\check{T} = 2\tau$} whenever a support exchange occurs,
and decrement it by \mbox{$\rho = 0.01$s} in every iteration of the control loop.

\subsection{Predictive Filter}
\label{chap:predictivefilter}

\begin{figure}[h]
\centering 
\includegraphics[height=2cm]{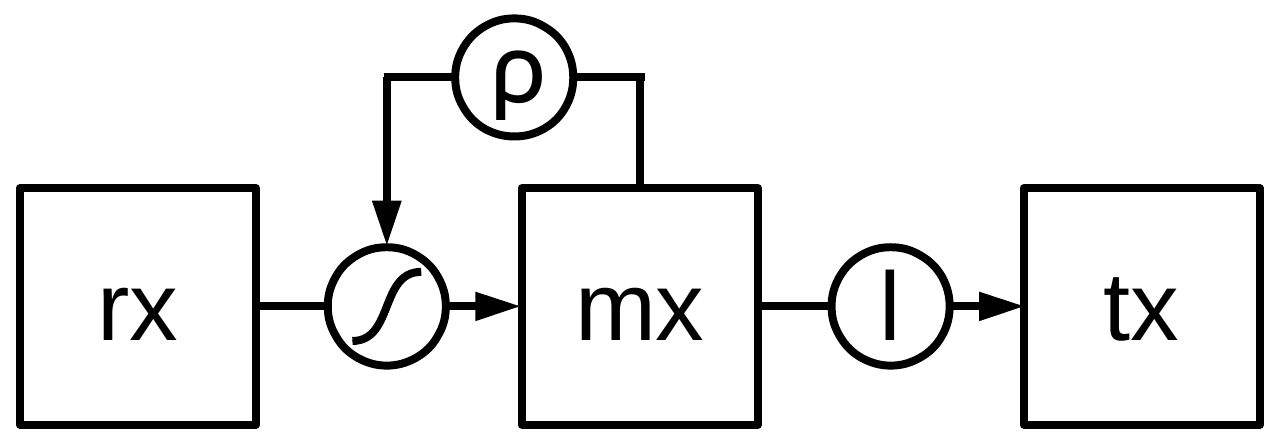}
\caption[Predictive Filter]{Predictive Filter. This filter removes noise
by blending between a measured \acl{CoM} state $\boldsymbol{rx}$ and an expected 
state $\boldsymbol{mx}$. The filter also predicts a short-term future state 
$\boldsymbol{tx}$ to compensate for latency.}
\label{fig:predictivefilter}
\end{figure}

In a real hardware environment, the sensor noise and the latency in the control
loop can have a significantly derogating effect on the performance of the system. 
We use a Predictive Filter as shown in Figure~\ref{fig:predictivefilter} to remove noise
from the \ac{CoM} state estimate and to compensate for the latency by making a
short-term prediction with the \ac{LIPM}. The three building blocks of the
filter are denoted $\boldsymbol{rx}$, $\boldsymbol{mx}$, and $\boldsymbol{tx}$.
The $\boldsymbol{rx}$ block is the \ac{CoM} state computed from the raw sensor
input by the State Estimation. The second building block named $\boldsymbol{mx}$
is a model state. In every iteration of the main control loop, the
$\boldsymbol{mx}$ state is simulated using the \ac{LIPM} by the time period $\rho = 0.01\,$s of 
one iteration of the control loop. The simulated state is then linearly
interpolated with the $\boldsymbol{rx}$ state using a blending factor $b \in
[0,1]$ 
\begin{equation}
\boldsymbol{mx} = b\,\boldsymbol{rx} + (1-b) \boldsymbol{mx}.
\end{equation}
In this manner, the $\boldsymbol{rx}$-$\boldsymbol{mx}$ loop forms a noise
filter that blends between an expected state according to the \ac{LIPM} and a
raw input state estimated from the sensor input. The $\boldsymbol{tx}$ block
contains the $\boldsymbol{mx}$ state predicted by the latency $l = 0.054\,$s, again using 
the \ac{LIPM} equations. It is the $\boldsymbol{tx}$ \ac{CoM} state
that is presented to the \mbox{\textsc{BalanceControl}} computation step of the 
Footstep Control algorithm~\ref{alg:footstepcontrol}. Effectively, the footstep
controller does not compute the step parameters for the state that was last
measured, but for the state the robot is estimated to be in by the time
the motors execute the commands.

The blending factor $b = f_s f_d$
is the product of two noise suppression functions $f_s$ and $f_d$. The step noise suppression function
\begin{align}
f_s &= 1 - \exp\left( - \frac{\max\{t_s - \epsilon, 0\}^2}{2\epsilon^2}\right)
\end{align}
is zero for a short time
after the support exchange, and rises smoothly as $t_s$---the time
since the last support exchange---increases with time. 
$\epsilon = 0.07$ tunes the duration of the step noise suppression. The second function
\begin{align}
f_d &= k \| \boldsymbol{rx} - \boldsymbol{mx} \|
\end{align}
suppresses blending when the $\boldsymbol{rx}$ and $\boldsymbol{mx}$
states are close. $k = 0.5$ is a gain. This way, the gait controller has a tendency towards
following an open-loop trajectory when the state of the system develops as expected and avoids the
possibly destabilizing effects of sensor noise.

\subsection{Balance Control}
\label{chap:balancecontrol}

\begin{figure}[h]
\centering 
\includegraphics[height=3cm]{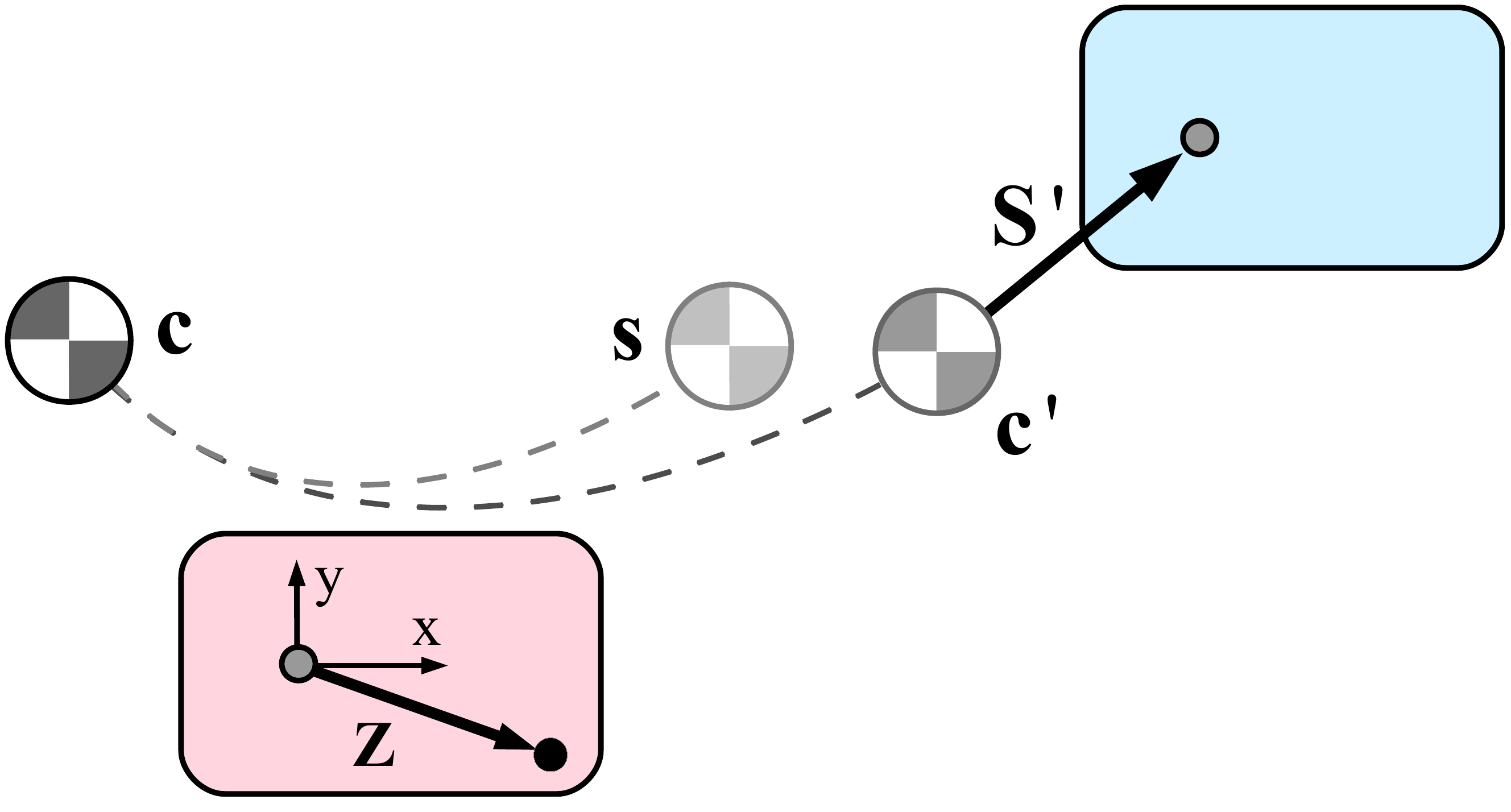}
\caption[The Concept of the Balance Controller]{The balance controller computes
a \acl{ZMP} offset $\boldsymbol{Z}$ that steers the \acl{CoM}
$\boldsymbol{c}$ towards the nominal support exchange state $\boldsymbol{s}$.
The \acl{ZMP} is not always effective in reaching the nominal
state, but the achievable end-of-step state $\boldsymbol{c}^{\prime}$ can be predicted.
The location of the next footstep $\boldsymbol{S}^{\prime}$ is computed with
respect to the achievable state $\boldsymbol{c}^{\prime}$.}
\label{fig:balancecontrol}
\end{figure}

The next computation step of the Footstep Control
algorithm~\ref{alg:footstepcontrol} is the
\mbox{$(\boldsymbol{Z}, \boldsymbol{A}, T)$ = \textsc{BalanceControl}($\boldsymbol{s}, \boldsymbol{c}, \lambda$)}
function in line 4.
Given the nominal target state $\boldsymbol{s}$, the filtered and
latency-compensated \ac{CoM} state $\boldsymbol{c}$, and the sign $\lambda$ of
the support leg, 
the balance control function computes the \ac{ZMP} offset
$\boldsymbol{Z}$, the swing amplitude $\boldsymbol{A}$, and the step time $T$
parameters that make the robot track the commanded velocity
$\boldsymbol{\check{V}}$ while maintaining balance.

The concept of the balance controller is illustrated in
Figure~\ref{fig:balancecontrol}. 
The balance controller computes a \ac{ZMP} offset
$\boldsymbol{Z}$ that steers the \ac{CoM} $\boldsymbol{c}$ towards the nominal location
$\boldsymbol{s}$. Since the \ac{ZMP} is physically bound to
remain inside the support polygon, it has only limited effect and the
nominal state is not guaranteed to be reached, but after the time $T$ for the support
exchange is chosen, the balance controller predicts the
achievable end-of-step state $\boldsymbol{c}^{\prime}$ and uses it to compute
the step coordinates $\boldsymbol{S}^{\prime}$ expressed relative to $\boldsymbol{c}^{\prime}$. 
Finally, the swing amplitude $\boldsymbol{A}$ is derived from the step coordinates $\boldsymbol{S}^{\prime}$. For the computation of the aforementioned
quantities, the following formulae are derived from the \ac{LIPM}.

\subsubsection{Lateral Zero Moment Point Offset}

By our design, the role of the lateral \ac{ZMP} $Z_y$ is to help maintaining the
nominal frequency of the lateral \ac{CoM} oscillation. We achieve this by computing
the lateral \ac{ZMP} $Z_y$ such that it accelerates the \ac{CoM} to reach the lateral
support exchange location $s_y$ at the nominal step time $\check{T}$. Using the current
\ac{CoM} state $\boldsymbol{c} = \left(c_x, \dot{c}_x, c_y,
\dot{c}_y\right)$, the nominal support exchange location $\boldsymbol{s} =
\left(s_x, \dot{s}_x, s_y, \dot{s}_y \right)$, and the nominal step time $\check{T}$
in Eq.~(\ref{eq:lipstatex}), we set $s_y = x(\check{T}, c_y-Z_y, \dot{c}_y)+Z_y$ and solve for $Z_y$. We obtain 
\begin{equation}
Z_y = \frac{ c_y \cosh(C \check{T}) + \frac{\dot{c}_y}{C} \sinh(C \check{T}) - s_y}{\cosh(C \check{T})-1}. 
\label{eq:zmpy}
\end{equation}
$Z_y$ then has to be bounded to a reasonable range $[Z_y^{min}, Z_y^{max}]$, for
example the width of the foot. The target lateral velocity $\dot{s}_y$ at the
support exchange location is neglected, but a possible error in the lateral
velocity at the support exchange location is corrected later on by the choice of
the lateral step size.

\subsubsection{Step Time}

\begin{figure}[h]
\centering 
\subfloat[]{\includegraphics[height=2.4cm]{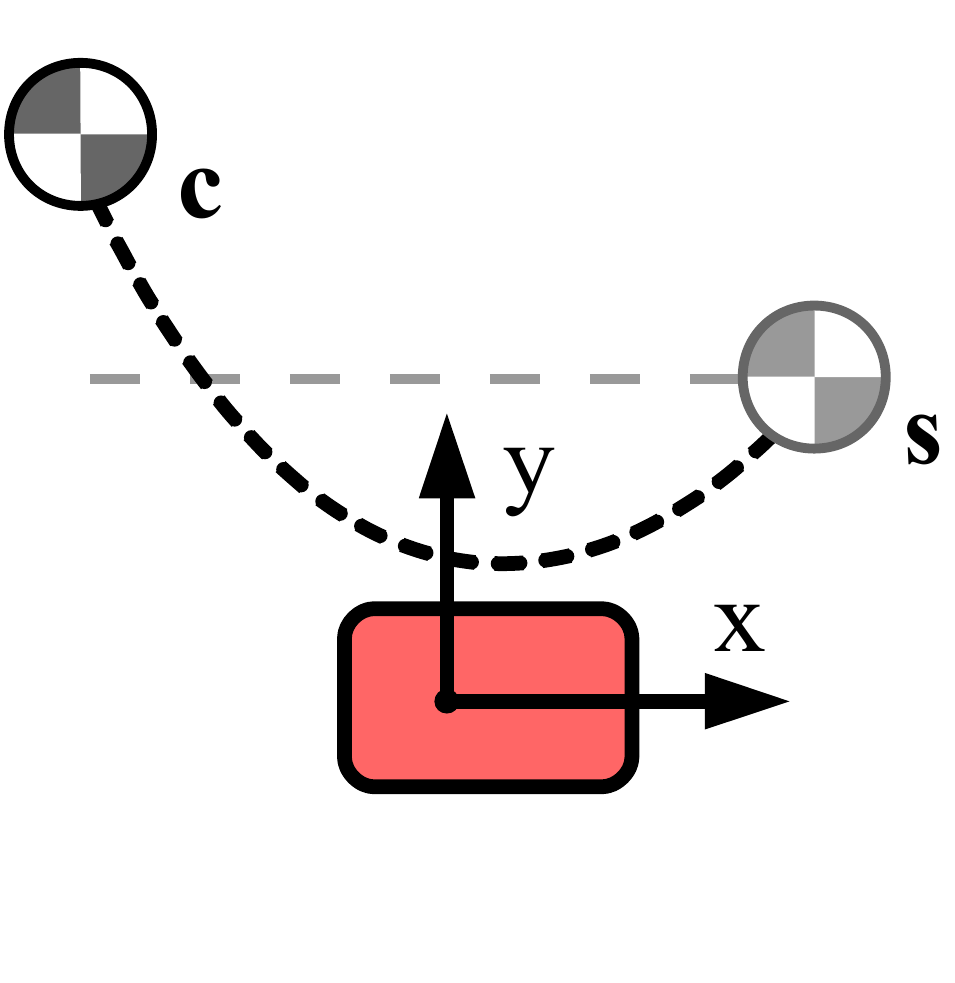}\label{fig:steptimea}}
\hspace{0.4cm}
\subfloat[]{\includegraphics[height=2.4cm]{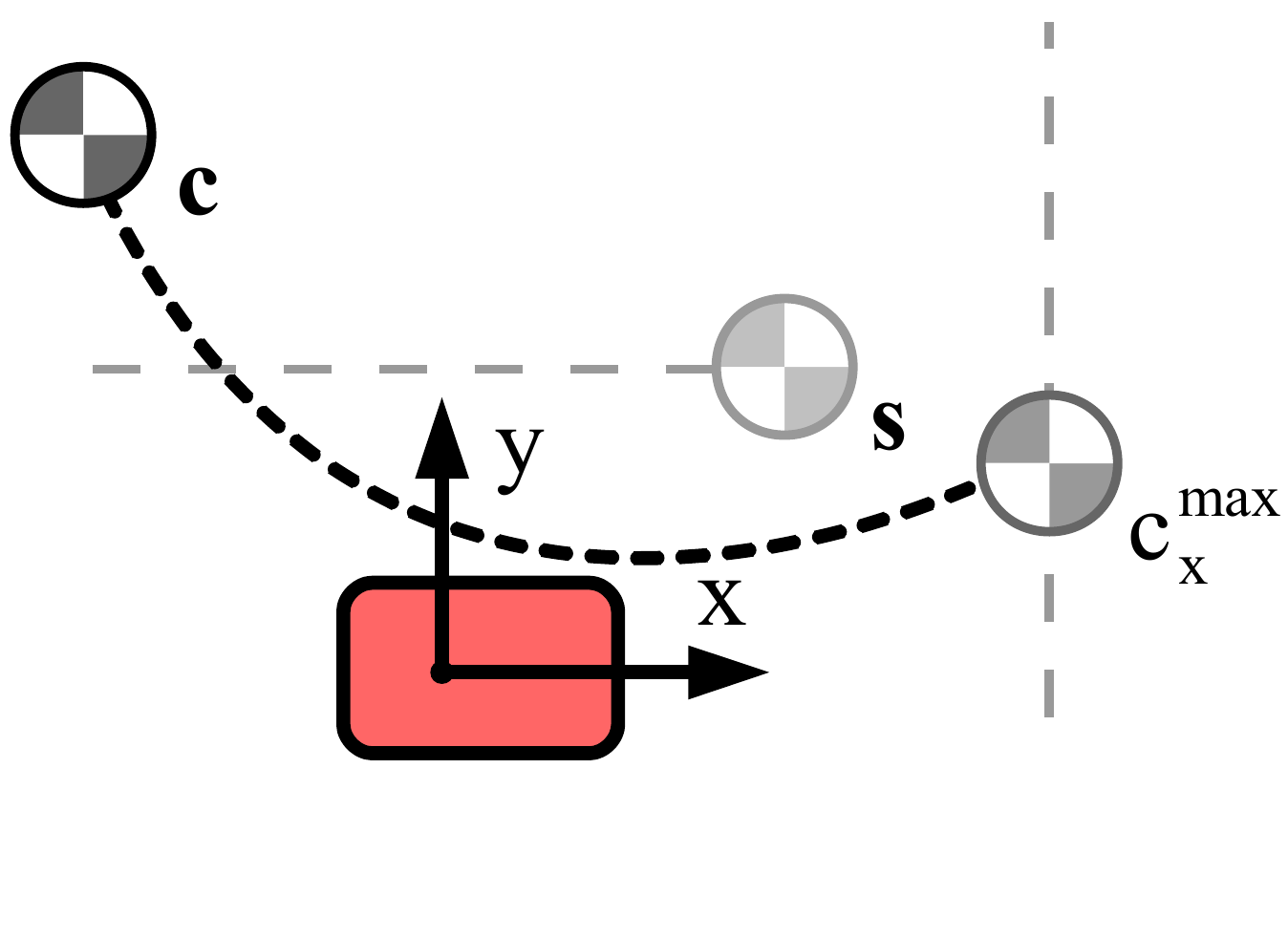}\label{fig:steptimeb}}
\hspace{0.2cm}
\subfloat[]{\includegraphics[height=2.4cm]{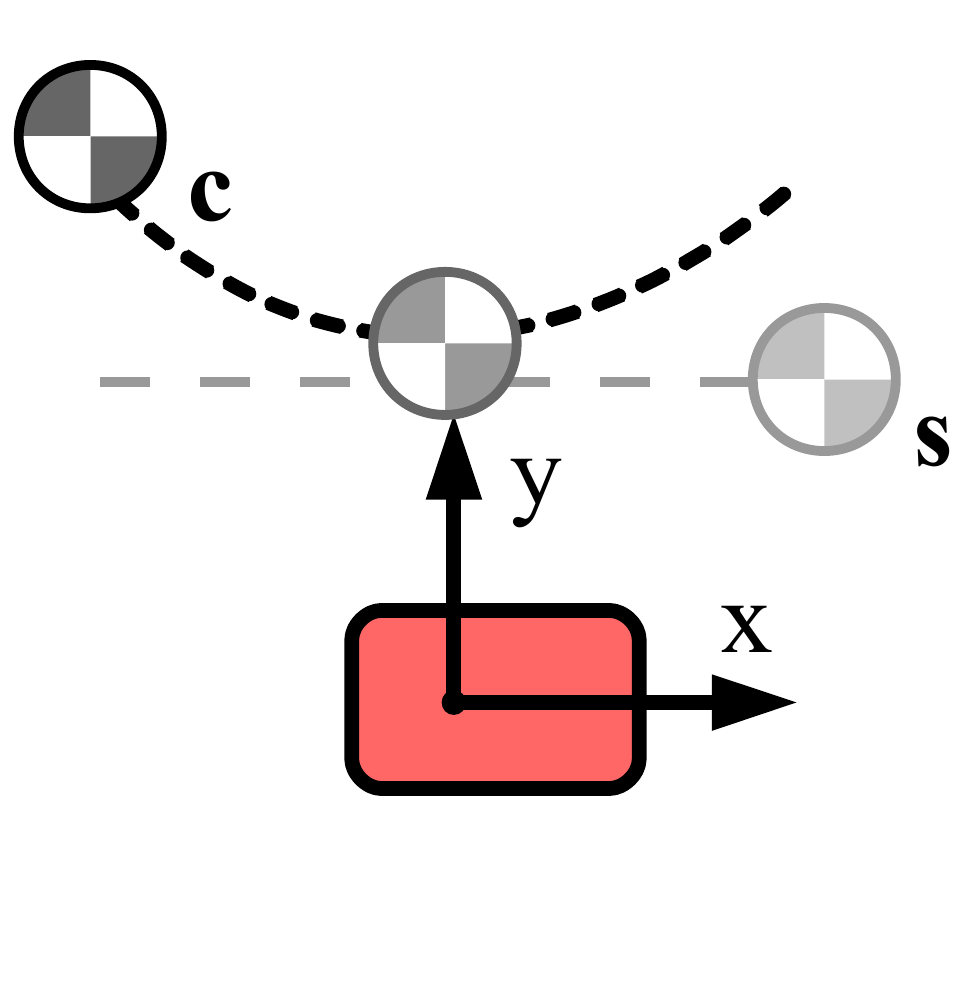}\label{fig:steptimec}}
\hspace{0.4cm}
\subfloat[]{\includegraphics[height=2.4cm]{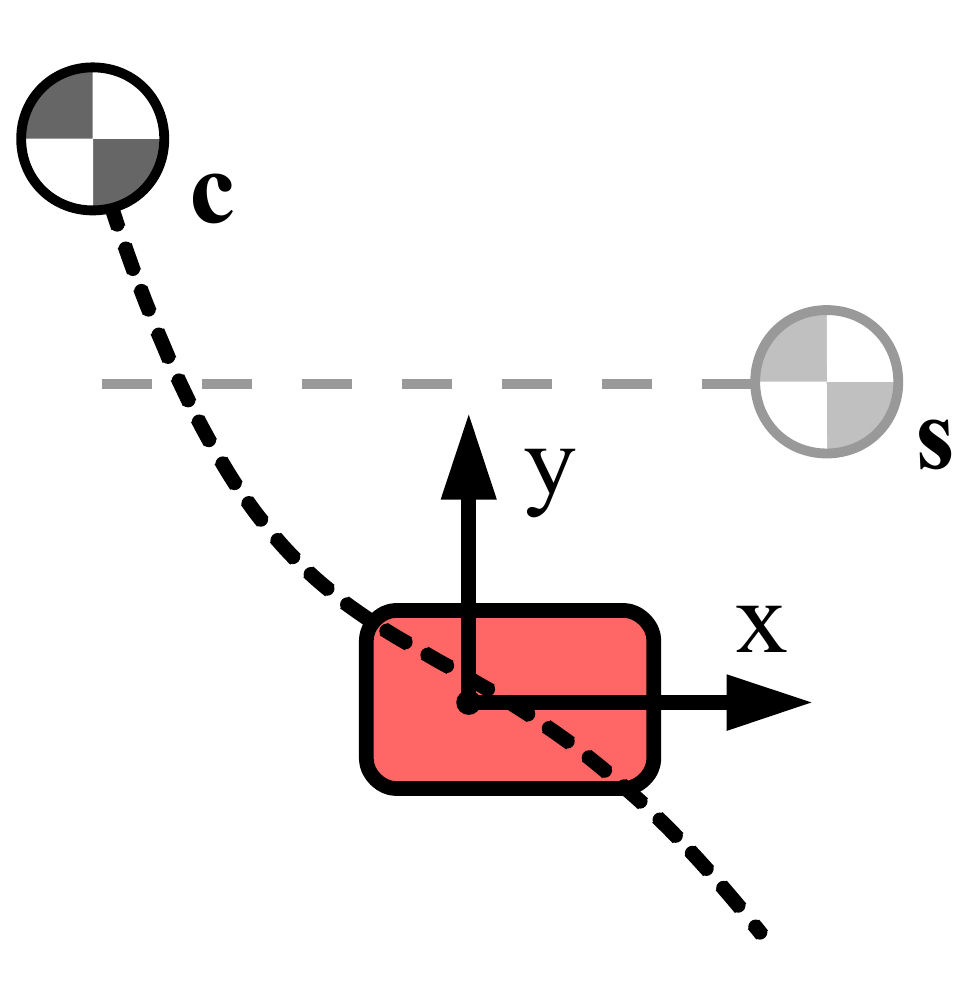}\label{fig:steptimed}}
\caption[Step Time Computation]{The balance controller
estimates the remaining time of the step as the time when (a) the \acl{CoM}
$\boldsymbol{c}$ reaches the lateral coordinate of the target location
$\boldsymbol{s}$. Special cases, such as (b) reaching the sagittal limit
$c_x^{max}$ first, (c) never reaching the lateral
coordinate, or (d) crossing the support foot, are handled explicitly.}
\label{fig:steptime}
\end{figure}

The next step parameter to compute is the step time $T$. Motivated by the
observed sensitivity of the lateral oscillation to disturbances as demonstrated in 
video\footnote{https://youtu.be/l9uvBD9zmsw}, we assume the lateral
oscillation to be the main determinant of the step time. The best
time for the support exchange is when the \ac{CoM} reaches the nominal lateral
support exchange location~$s_y$. In this position, the robot can be expected to
be upright and to have sufficient lateral momentum to transfer its weight to the
other leg. The ideal case is illustrated in Figure~\ref{fig:steptimea}. During a typical step, the
\ac{CoM} travels towards the support leg, changes direction at the apex, and eventually
reaches the nominal support exchange location~$s_y$. Using the
\ac{LIPM} time-of-location equation~(\ref{eq:liptimex}) including the influence of
the lateral \ac{ZMP}~$Z_y$, the time to reach $s_y$ is given by
\mbox{$T(s_y) = t_{pos}(s_y\!-\!Z_y, c_y\!-\!Z_y, \dot{c}_y)$}.
There are, however, special cases that must be considered.
Figure~\ref{fig:steptimeb} shows the
case, where a sagittal limit $c_x^{max}$ is reached before $s_y$, beyond which
an increase of the stride length would also compromise balance. The time to 
reach $c_x^{max}$ is given by $T(c_x^{max}) = t_{pos}(c_x^{max}, c_x, \dot{c}_x)$.
Figure~\ref{fig:steptimec} shows a case where the support exchange location
$s_y$ is never reached. A strong disturbance can cause the \ac{CoM} to
never come across the support exchange coordinate. Situations where the support
exchange location has been crossed in the past without a step having occurred
also belong to this category. Case (c) can be detected when
$T(s_y)$ does not compute a positive value. 
In that case, if a positive time $t_{vel}(0, c_y-Z_y, \dot{c}_y)$
can be determined using the \ac{LIPM} time-of-velocity
equation (\ref{eq:liptimevx}) with a target velocity of zero, then an irregular
lateral apex is still to be encountered in the future. The irregular apex is
the closest point to the lateral support exchange location, and the time to
reach this apex can be used as a sensible step time. Otherwise, we set the step
time to zero and the balance controller recommends an immediate step, which
drives the step motion generator at its maximum permitted frequency towards the
next support transition. Finally, Figure~\ref{fig:steptimed} shows the critical
case where the \ac{CoM} is estimated to tip over the support foot, indicated by a
positive lateral orbital energy $E(c_y,\dot{c}_y)$
(Eq.~\ref{eq:lipenergy}). In this case, we use a large constant step time of $T =
2$ seconds to slow the stepping motion down and hope that the \ac{CoM} will
return after all. If the robot does tip over, a recovery step cannot reasonably be 
taken and the robot will fall.

All cases considered, the step time parameter $T$ is given by
\begin{equation}
T = 
\begin{cases}
T(c_x^{max}), & \mbox{if } T(c_x^{max}) < T(s_y),\\
T(s_y), & \mbox{if } T(s_y) > 0 \wedge T(s_y) < \infty,\\
t_{vel}(0, c_y\!-\!z_y, \dot{c}_y), & \mbox{if } t_{vel}(0, c_y\!-\!z_y, \dot{c}_y) > 0 \wedge T(s_y) = \infty,\\ 
2, & \mbox{if } E(c_y,\dot{c}_y) > 0,\\
0, & \mbox{otherwise}.\\
\end{cases}
\end{equation}
The step parameters computed in the following depend on the
step time T.

\subsubsection{Sagittal Zero Moment Point Offset}

In the sagittal direction, we compute the \ac{ZMP} such that the sagittal support
exchange location $s_x$ is reached at the step time $T$. We compute the sagittal
\ac{ZMP} offset as
\begin{align}
Z_x &= \frac{c_x \cosh(C T) + \frac{\dot{c}_x}{C} \sinh(C T) - s_x}{\cosh(C T) \label{eq:zmpx}
-1}
\end{align}
and bound it to the range $[Z_x^{min}, Z_x^{max}]$.
Note that in the sagittal direction, we aim for the \ac{CoM} to arrive at the sagittal
support exchange location at the predicted step time $T$ unlike in the lateral
direction, where we aimed for the nominal step time $\check{T}$.

\subsubsection{Footstep Location}

The choice of the next footstep coordinates has a
strong influence on the future trajectory of the \ac{CoM}. Our concept to
determine a suitable footstep location is based on prediction. Given the current
\ac{CoM} state $\boldsymbol{c}$ and the bounded \ac{ZMP} offset $\boldsymbol{Z}$ computed in equations (\ref{eq:zmpy}) and (\ref{eq:zmpx}), 
we estimate the achievable end-of-step state 
\mbox{$\boldsymbol{c^{\prime}}$ = \textsc{LipmPredict($\boldsymbol{c}$, $\boldsymbol{Z}$, $T$)}} (Eq.~\ref{eq:limppredict})
that will be reached by the already chosen step time $T$. After a disturbance, the
achievable state can significantly deviate from the nominal state
$\boldsymbol{s}$.

In the following, we compute the sagittal and lateral coordinates of the
footstep \mbox{$\boldsymbol{S}^{\prime} = \left(S^{\prime}_x,
S^{\prime}_y\right)$} expressed relative to the coordinates of
the predicted end-of-step state $\boldsymbol{c}^{\prime}$. Due to their
conceptually distinct behavior, we use different strategies for the sagittal and
the lateral directions. In the sagittal direction, we simply use our step symmetry
assumption where the \ac{CoM} is in the center between the feet in the moment of
the support exchange and set 
\begin{equation}
S^{\prime}_x = c^{\prime}_x.
\end{equation}
The lateral step size $S^{\prime}_y$ is computed such that the \ac{CoM} will
pass the apex of the next step at a distance $\alpha$, \ie, the
orbital energy $E(S^{\prime}_y, \dot{c}^{\prime}_y)$ (Eq.~(\ref{eq:lipenergy})) right after
the support exchange should
equal the constant energy level of the lateral step apex $E(\alpha, 0)$.
Solving the equation $E(S^{\prime}_y, \dot{c}^{\prime}_y) = E(\alpha, 0)$ for
$S^{\prime}_y$ yields the lateral step size
\begin{equation}
S^{\prime}_y = \lambda \sqrt{\frac{\dot{c}^{\prime 2}_y}{C^2} + \alpha^2},
\end{equation}
where $\lambda \in \{-1,1\}$ is the sign of the support leg prior to the step. 

The conversion of the step size to the swing amplitude $\boldsymbol{A}$
is straight forward.
\begin{align}
A_x &= \frac{S^{\prime}_x}{2}\sigma,\\
A_y &= \frac{\frac{S^{\prime}_y}{2} - \delta}{\omega - \delta},\\
A_{\psi} &= \check{V}_{\psi}.
\end{align}
The rotational swing amplitude $A_{\psi}$ is not considered to be relevant for
balance and is simply passed through from the commanded rotational velocity.

The now complete step parameters $\left(\boldsymbol{A}, T\right)$ are passed on
to the Motion Generator module to command the robot to step to the computed
location at the right time. The \ac{ZMP} is not passed on to the Motion
Generator. As the swing vector $\boldsymbol{A}$ and the step time $T$ are
implicit results of the computed \ac{ZMP} offset, and the two quantities are 
physically linked, the execution of the commanded step should
in theory place the \ac{ZMP} roughly in the right
location under the foot.

\section{Experiments}
\label{chap:analyticexperiments}

To demonstrate the effectiveness of the introduced gait controller, we provide
video material\footnote{https://youtu.be/yOQql5eSjn8} where the controllability
and the disturbance rejection capabilities of the robot can be seen. We also
present a systematic evaluation of a large number of pushes that our
robot NimbRo-OP2 was subjected to. 

\subsection{NimbRo-OP2 Robot}
\label{chap:therobots}

\begin{figure}[h]
\centering
\subfloat[The Robot]{\includegraphics[height=4.0cm]{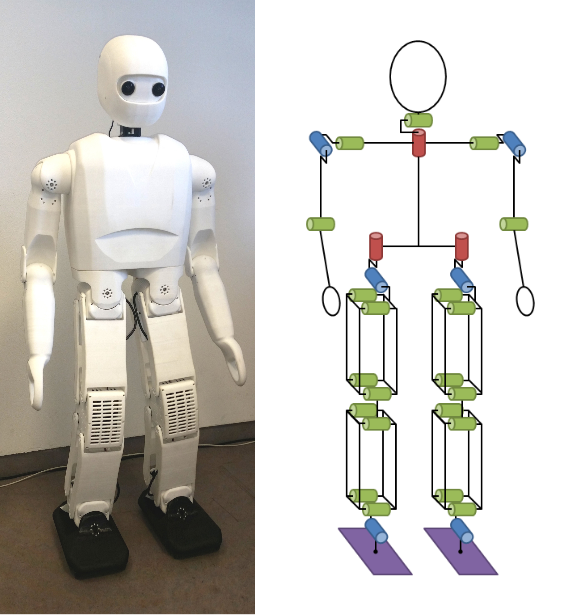}\label{fig:robota}}
\hspace{1.0cm}
\subfloat[Compliant Actuation]{\includegraphics[height=4.0cm]{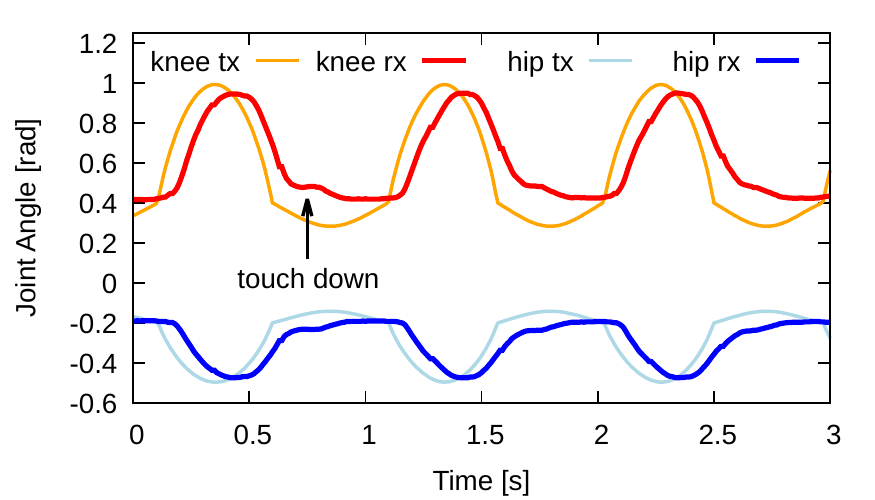}\label{fig:robotb}}
\caption[The Robot]{(a) The NimbRo-OP2 robot. (b) Commanded (tx) and measured (rx) 
joint angles of the right knee and the right hip during walking. The 
latency and the deviation of the actuators from the commanded position 
are quite large, but the shock of the touch-down at the end of the step is 
automatically absorbed.}
\label{fig:robot}
\end{figure}
\vspace{-10pt}

The NimbRo-OP2 robot \citep{NimbRo-OP2} that we used to validate our bipedal gait controller is an
open-source hardware platform. The robot is 134.5$\,$cm tall and weighs
17.5$\,$kg. It has a 3D printed Polyamide exoskeleton and uses Robotis Dynamixel
MX-106R servo motors as actuators. It features an Intel NUC PC with an
Intel Core i7-7567U 3.5-4.0\,GHz CPU and 4\,GB RAM for onboard processing. 
The legs of the robot were constructed with a parallel kinematics structure
as shown in Figure~\ref{fig:robota}. The parallel linkages in the
thigh and in the shank mechanically force the knee joint to stay parallel to
the trunk, and the foot plate to remain parallel to the knee joint. This
construction lacks the ankle pitch degree of freedom, but provides
additional passive stability. Using the walk described in this work, the NimbRo-OP2 robot won the RoboCup soccer competitions in 2019 where it used capture steps to maintain its balance during soccer games\footnote{https://youtu.be/ITe-seb4PsA} and excelled in the 
Technical Challenges\footnote{https://youtu.be/4aVTt2iSry4}.

The actuators support a compliant 
setting where the position controller can be configured with a low gain. This results
in a relatively compliant actuator behavior with a soft feel at the cost of imprecise
position tracking. Figure~\ref{fig:robotb} shows the commanded 
(tx) and measured (rx)
motion trajectories of the knee and hip joints in the right leg during walking. 
The latency and the deviations between the commanded and the received positions 
are quite evident, but also the automatic absorption of the floor impact can be seen.

\subsection{Walking Push Recovery Experiment}
\begin{figure}[b]
\centering
{\includegraphics[width=0.49\textwidth]{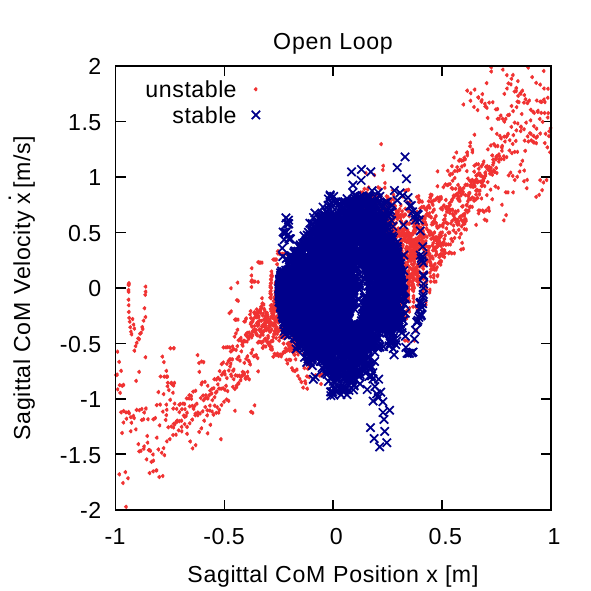}
\includegraphics[width=0.49\textwidth]{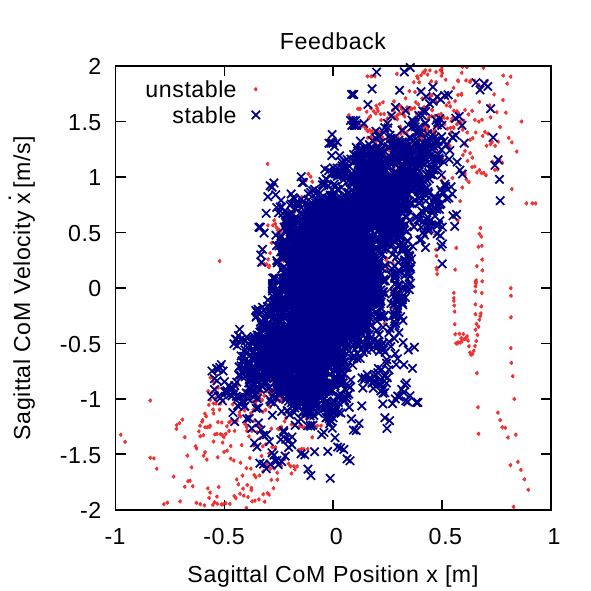}}\\
{\includegraphics[width=0.49\textwidth]{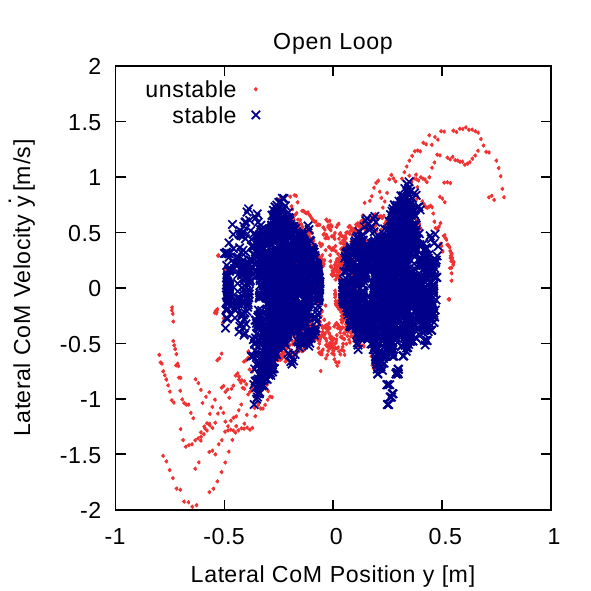}
\includegraphics[width=0.49\textwidth]{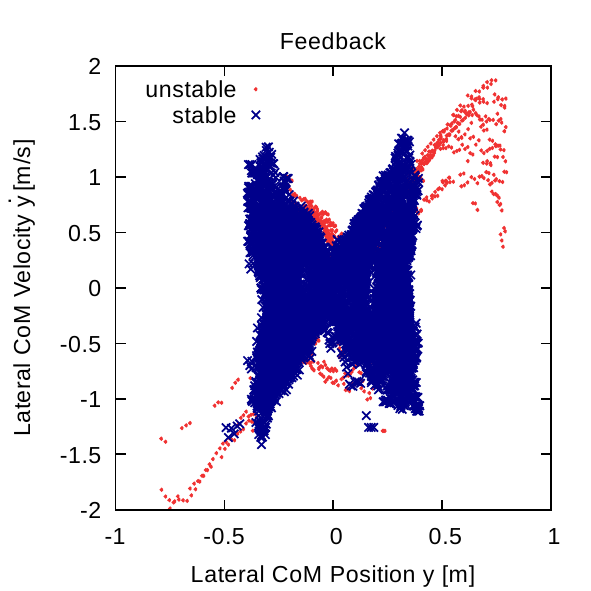}}\\
\caption{Stability analysis. Sagittal (top) and lateral (bottom) phase plots with data collected during an open loop (left) and a closed loop (right) push-recovery experiment. Stable regions are marked in dark blue color, unstable regions where the robot falls are marked in light red.}
\label{fig:pushsynchronizedopenloop} 
\end{figure}
In a push recovery experiment, we explored the effectiveness of the
analytic capture step controller by subjecting the robot to a large number of
pushes from the front, from the back, from the left, and from the right. The
pushes were applied by hand with varying strength. We made sure to include
strong enough pushes in each direction that were beyond the capabilities of the
capture step controller in order to explore its limitations. In total, we
applied 58 pushes while the robot was walking in open-loop mode without capture
step control out of which the robot fell 22 times. Then we applied another 97
pushes to the robot with the capture step controller enabled and the robot fell
18 times. Note that the falls to number of pushes ratio is not a good indicator
of stability as the strength of the pushes varied between the open-loop and the
closed-loop sessions. Stronger pushes were applied to drive the active
controller to its limits. It is rather an indicator for the fact that the
experiments were performed in a range of disturbances that includes pushes
strong enough to push the robot over.

Figure~\ref{fig:pushsynchronizedopenloop}
shows an evaluation in the sagittal and the lateral phase spaces of
the center of mass. 
We can observe
stable regions where the robot would not fall even when walking
with the open-loop motion generator, but these regions are much larger when
the capture step controller is active. The existence of a stable region of the open-loop step generator is best
explained by the non-zero size of the feet and the stiff ankle joint. Light pushes
drive the center of mass away from the center position, but as long as the
center of mass does not leave the edge of the foot, it has a good chance of
returning to the gait limit cycle.

\begin{figure}[b]
\centering
\includegraphics[height=3.5cm]{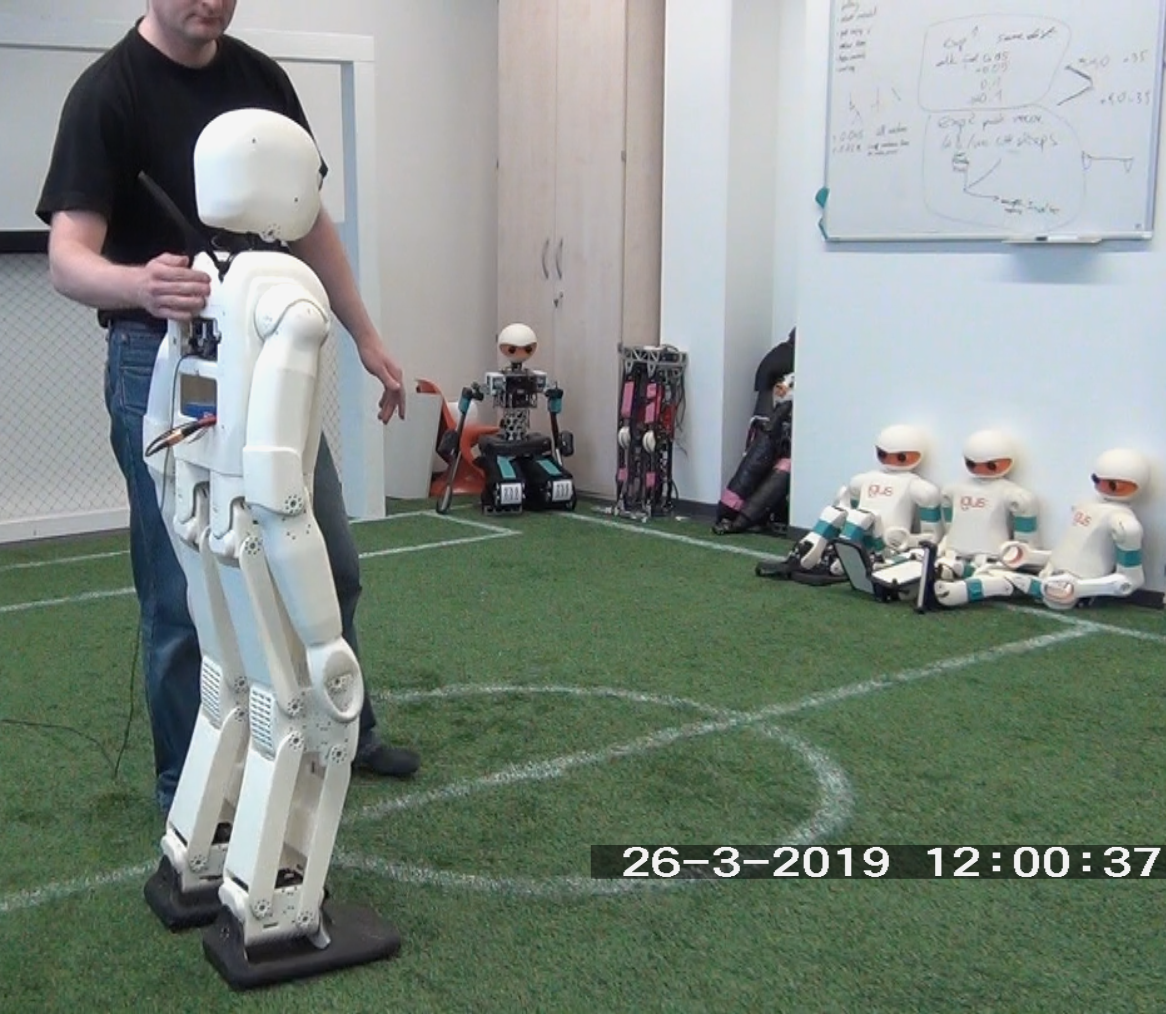}
\includegraphics[height=3.5cm]{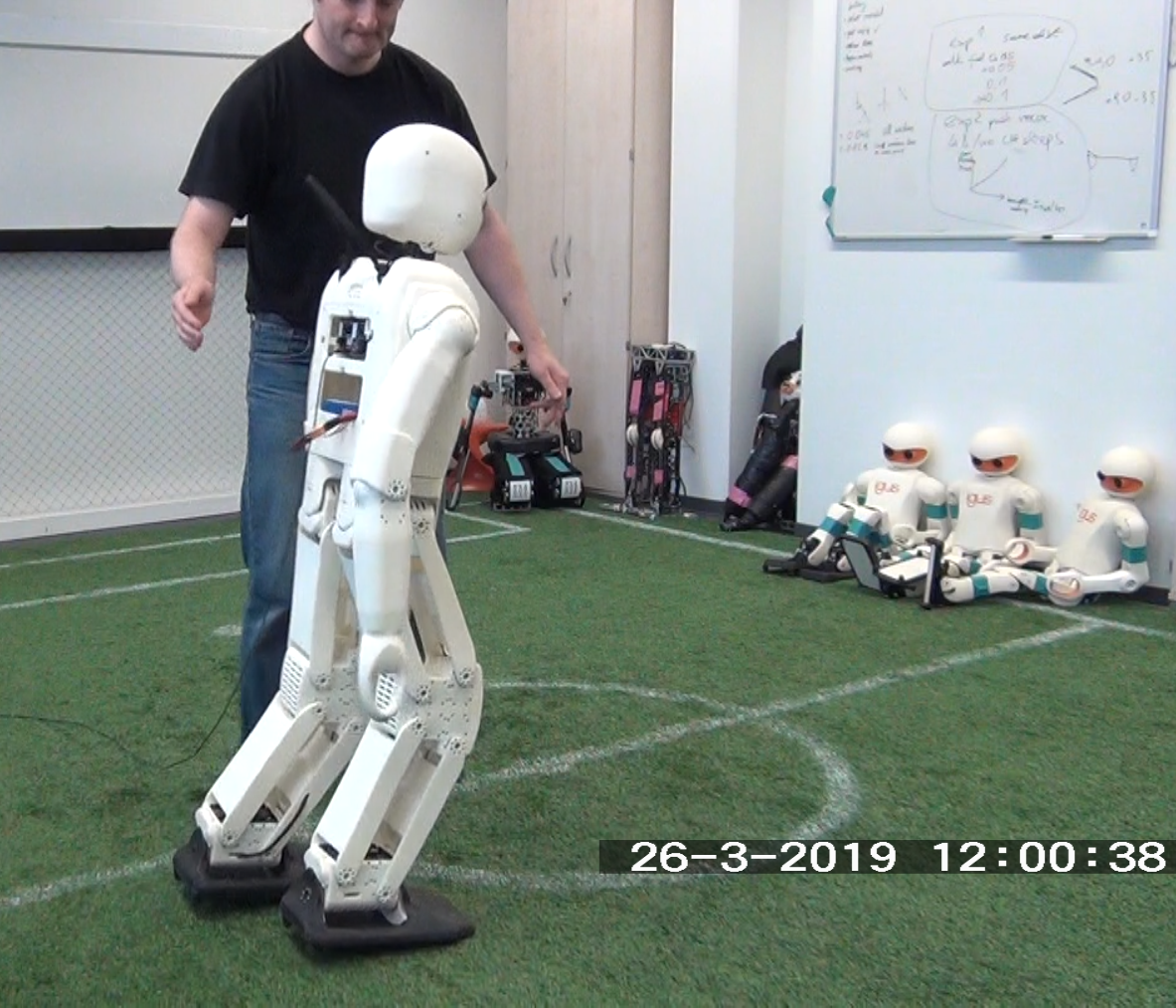}
\includegraphics[height=3.5cm]{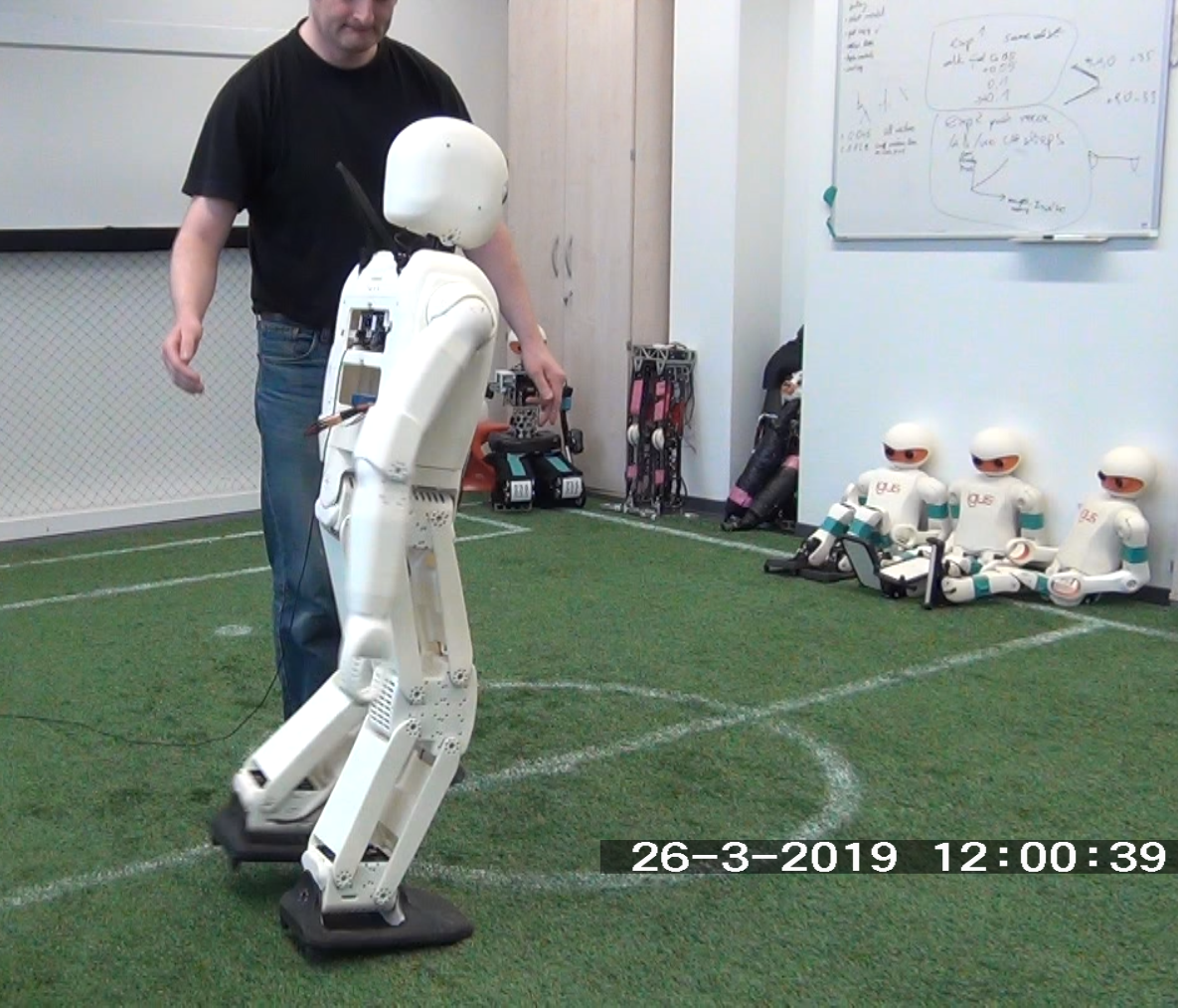}
\includegraphics[height=3.5cm]{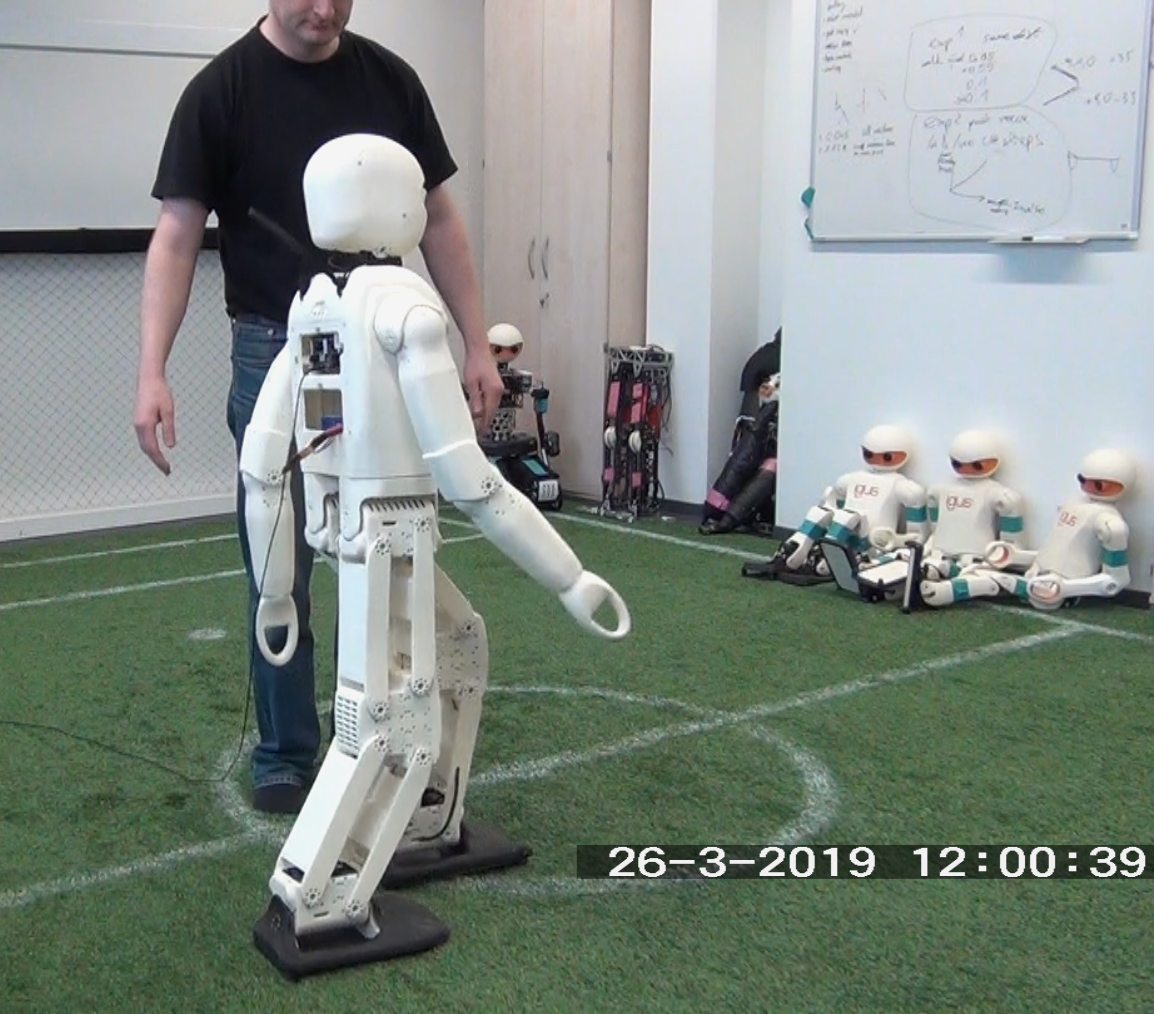}
\includegraphics[height=3.5cm]{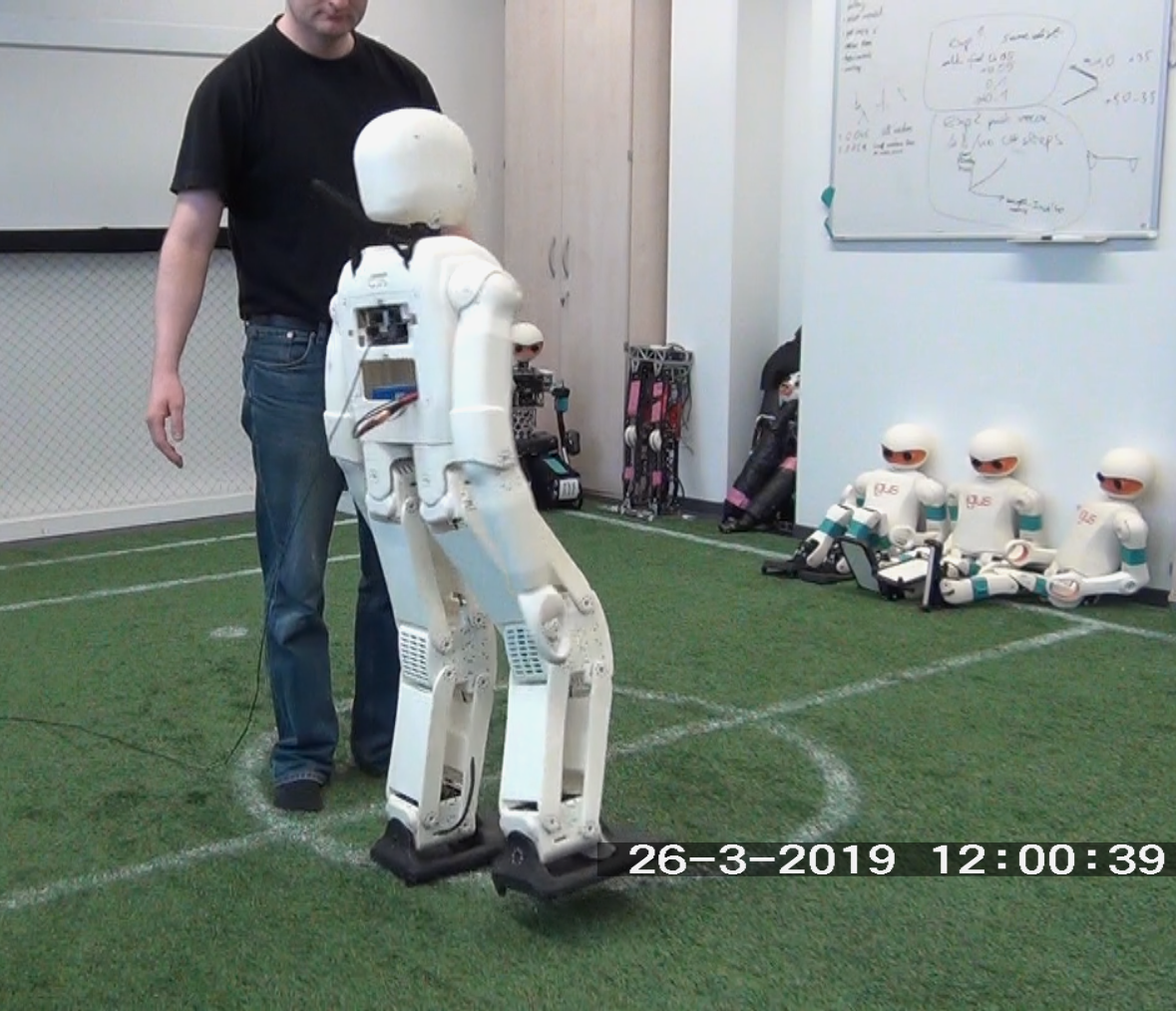}
\includegraphics[height=3.5cm]{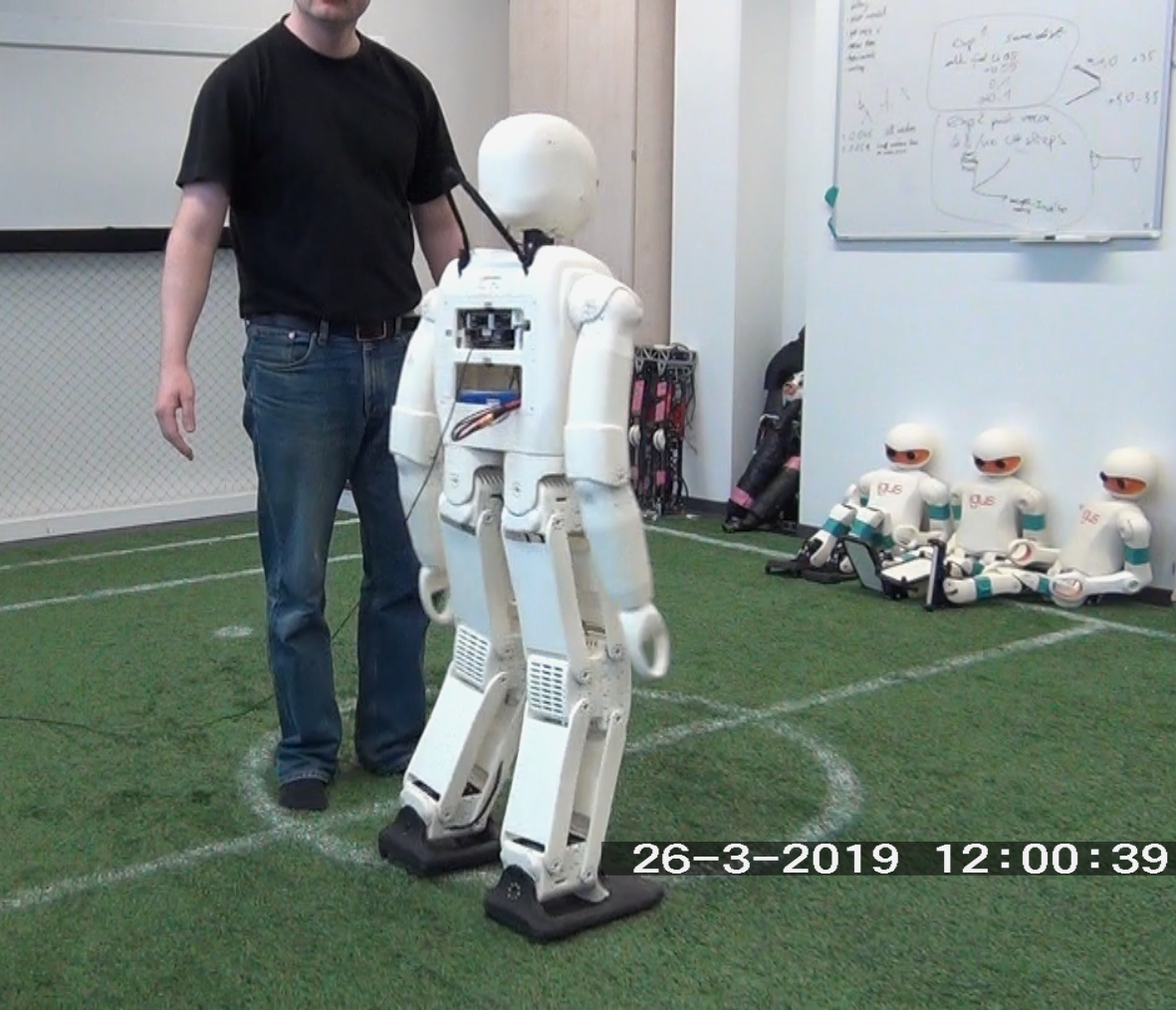}
\caption{From top left to bottom right: NimbRo-OP2 performing capture steps after a push.}
\label{fig:pushexperiment2}
\vspace{-10pt}
\end{figure}

\begin{figure}[t]
\centering
\includegraphics[width=\textwidth]{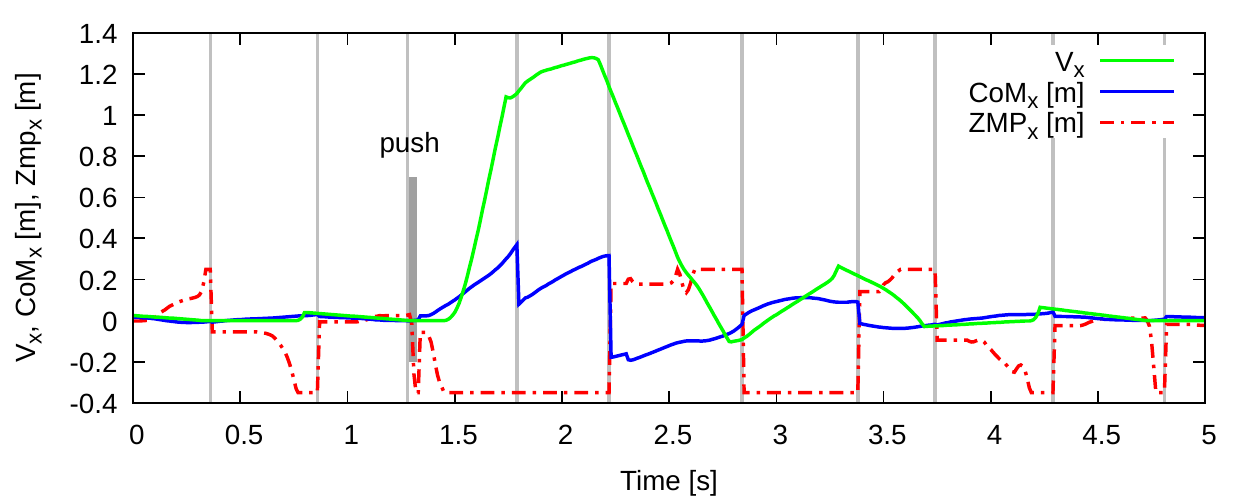}
\caption{\acl{CoM}, \acl{ZMP} and swing amplitude data recorded with NimbRo-OP2 after a sagittal push. Support exchange moments are indicated by thin vertical lines. The thick vertical bar marks the time of the push.}
\label{fig:pushexperiment}
\vspace{-10pt}
\end{figure}

A photo strip of a selected push from the back is shown in
Figure~\ref{fig:pushexperiment2}. The sagittal \ac{ZMP}, the
sagittal \ac{CoM} position, and the swig amplitude $A_x$ during the recovery of this
push are shown in Figure~\ref{fig:pushexperiment}. The push happened in an
unlucky moment shortly after the support exchange where the noise
suppression ignores the sensor input for a short while. We can observe how the
\ac{ZMP} moves into the heel after the push to accelerate the robot forward and
it remains there for two steps while the robot keeps accelerating, as shown by
the rising step amplitude $A_x$. The first step is not enough to capture the
escaping \ac{CoM} shown by the blue line. Only with the second step the robot
manages to move the center of mass behind the foot. During the third step, the
\ac{ZMP} is in the toe and the robot decelerates. It takes a few more steps to
settle the \ac{CoM} close enough to zero where no more step size modification is
needed.

\subsection{Dynaped Videos}

In earlier work, we evaluated the same capture step controller with our robot Dynaped \citep{Missura:WalkingWithCaptureSteps}. The video\footnote{http://youtu.be/PoTBWV1mOlY} shows reliable and
controllable walking skills with strong disturbance rejection capabilities. Dynaped was not only disturbed by pushes during walking, but also by
placing a hand under its feet, and by forcing collisions with a static obstacle. In the experiment shown in video\footnote{http://youtu.be/GU53yomxrxE}, we mounted smaller feet of human-like
proportions on Dynaped, and after refitting the parameters, we reproduced omnidirectional walking capabilities of similar quality or even better than before. We also
managed to produce a few quite extreme cases of push recovery, as shown towards
the end of the video.


\section{Conclusion}
\label{chap:conclusion}

We introduced a new approach to robust bipedal walking with
push recovery capabilities. The core concept of the gait controller is
to use a CPG to generate open-loop stepping motions that can be controlled
in terms of step size and timing. Using this control interface allows the 
implementation of a low-dimensional balance controller that is derived
analytically with the help of a \acl{LIPM} that was specifically fitted to the open-loop motion of the robot.

In spite of the imprecision and the latency caused by the compliant setting of the
actuators, we are able to demonstrate robust and controllable omnidirectional
walking on a real robot with strong push recovery capabilities. We are able to
accomplish this without using a precise dynamic model of the robot, without
detecting foot contact, and without means of measuring or enforcing the
model-suggested location of the \acl{ZMP}. We have substantiated this claim
with systematic statistical experiments and video examples. 

The architecture of the Capture Step Framework gives rise to potential beyond
the concepts that have been explored so far. The manageable level of complexity
leaves sufficient room to be extended with additional functionalities such as
balance-restoring actions using the trunk or the arms. The separation of motion and balance should make machine learning tasks easier where either only the motion generator, or only the balance controller could be trained separately from each other. The reduced dimensionality of the learning tasks, and starting with a robot that can already walk, could help leveraging online learning algorithms to optimize the parameters of the motion patterns  and to learn to control the balance of the robot. Existing methods \citep{KatjaWalking} are often impractical to be applied on a real robot due to the number of repetitions they require. The fast computation times of the presented controller could be leveraged to implement balance-aware footstep planning.

\section*{Acknowledgements}

This work has been partially funded by grant BE 2556/13-1 of German Research Foundation (DFG). We thank Grzegorz Ficht for supporting the experiments with software for the NimbRo-OP2 robot.

\appendix


\bibliographystyle{plainnat}
\bibliography{Bibliography}	

\vspace*{2cm} 

\noindent%
\parbox{5truein}{
\begin{minipage}[b]{1truein}
\centerline{\includegraphics[width=2.6cm]{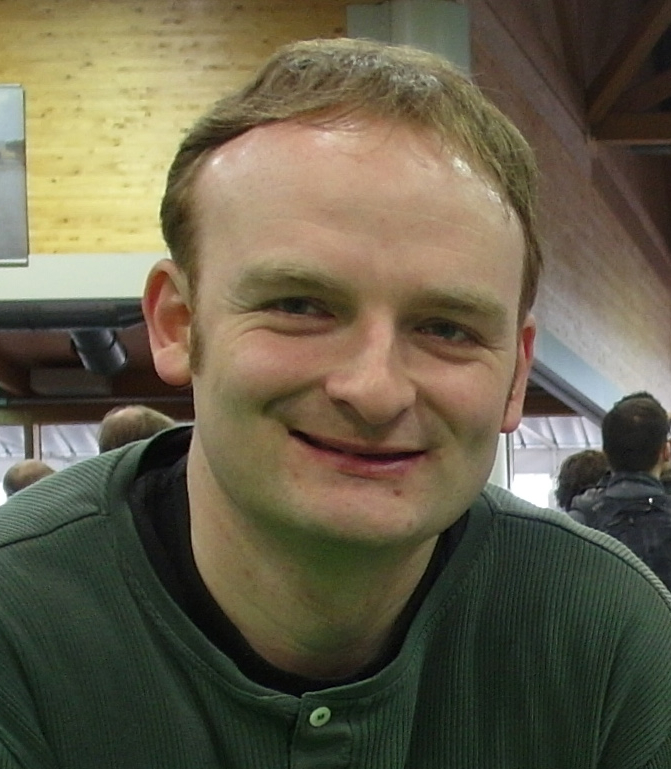}}
\end{minipage}
\hfill         
\begin{minipage}[b]{3.85truein}
{{\bf Marcell Missura} is a postdoc researcher and teaching assistant at the Humanoid Robots Lab of the University of Bonn in Germany. He obtained his PhD from the University of Bonn in computer science in 2016 for his work on bipedal walking with push recovery capabilities. Marcell has earned six world champion titles and three times the Louis Vuitton Best Humanoid Award in the international RoboCup competitions. \hfilneg}
\end{minipage} } 

\vspace*{3.9pt}  
\noindent
His research topics include motion planning in dynamic environments and robust bipedal walking for humanoid robots.

\vspace*{13pt}  
\noindent%
\parbox{5truein}{
\begin{minipage}[b]{1truein}
\centerline{\includegraphics[width=2.6cm]{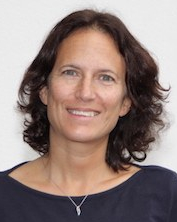}}
\end{minipage}
\hfill 
\begin{minipage}[b]{3.85truein}
{{\bf Maren Bennewitz} is professor for Computer Science at the University 
of Bonn, Germany, and head of the Humanoid Robots Lab.
She received her Ph.D. in Computer Science from the University of Freiburg
in 2004. Before she moved to Bonn in 2014, she was a Postdoc and assistant
professor at the University of Freiburg. The focus of her research lies on
robots acting in human environments. In the last few years, she has been
developing several innovative solutions for robotic systems co-existing
and\hfilneg}
\end{minipage} } 

\vspace*{4.8pt}  
\noindent
interacting with humans. Among them are techniques for efficient navigation
with humanoid and wheeled robots as well as for reliably detecting and
tracking humans from sensor data and analyzing their motions.

\vspace*{13pt}  
\noindent%
\parbox{5truein}{
\begin{minipage}[b]{1truein}
\centerline{\includegraphics[width=2.6cm]{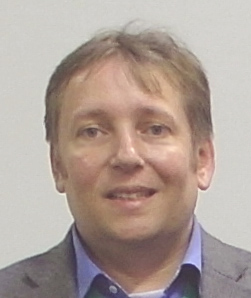}}
\end{minipage}
\hfill 
\begin{minipage}[b]{3.85truein}
{{\bf Sven Behnke} is full professor for Autonomous Intelligent Systems at Rheinische Friedrich-Wilhelms-Universit{\"a}t Bonn. He received his M.S. degree in Computer Science in 1997 from Martin-Luther-University, Halle-Wittenberg, and his Ph.D. degree in 2002 from Freie Universit{\"a}t Berlin. In 2003, he worked as a postdoc at the International Computer Science Institute, Berkeley. From 2004 to 2008, he headed the Hu--\hfilneg}
\end{minipage} } 

\vspace*{4.8pt}  
\noindent
manoid Robots group at Albert-Ludwigs-Universit{\"a}t, Freiburg. His research interests include cognitive robotics, computer vision, and machine learning.

\vfill\eject

\end{document}